\documentclass[final,3p,twocolumn,authoryear,sort&compress,times]{maia}
\journal{MAIA Master}

\usepackage{amsmath,amssymb}	
\usepackage{graphicx}
\usepackage{hhline}	
\usepackage{color}
\usepackage{natbib}
\usepackage{url}
\usepackage{amsmath}
\usepackage{amssymb}
\usepackage{booktabs, multirow} 
\usepackage{soul}
\usepackage[table]{xcolor} 
\usepackage{changepage,threeparttable} 
\usepackage{tabularx}
\usepackage{booktabs}

\definecolor{red}{RGB}{255,0,0}

\graphicspath{./figures/}

\begin{document}

\begin{frontmatter}

\title{SYNCS: Synthetic Data and Contrastive Self-Supervised Training for Central Sulcus Segmentation}

\author{Vladyslav Zalevskyi, Kristoffer Hougaard Madsen}
\address{Danish Research Centre for Magnetic Resonance (DRCMR), Hvidovre, Denmark}

\begin{abstract}
Bipolar disorder (BD) and schizophrenia (SZ) are severe mental disorders that have a significant impact on individuals and society. Early identification of risk markers for these diseases is crucial for understanding their progression and enabling preventive interventions. The Danish High Risk and Resilience Study (VIA) is a longitudinal cohort study that aims to gain insights into the early disease processes of SZ and BD, particularly in children with familial high risk (FHR). Understanding structural brain changes associated with these diseases during early stages is essential for effective interventions. The central sulcus (CS) is a prominent brain landmark related to brain regions involved in motor and sensory processing. Analyzing CS morphology can provide valuable insights into neurodevelopmental abnormalities in the FHR group. However, CS segmentation presents challenges due to its high morphological variability and complex shape, which are especially apparent in the adolescent cohort. This study explores two novel approaches for training robust and adaptable CS segmentation models that address these challenges. Firstly, we utilize synthetic data generation to model the morphological variability of the CS, adapting SynthSeg's generative model to our problem. Secondly, we employ self-supervised pre-training and multi-task learning to adjust the segmentation models to new subject cohorts by learning relevant feature representations of the cortex shape. These approaches aim to overcome limited data availability and enable reliable CS segmentation performance on diverse populations, removing the need for extensive and error-prone post- and pre-processing steps. By leveraging synthetic data and self-supervised learning, this research demonstrates how recent advancements in training robust and generalizable deep learning models can help overcome problems hindering the deployment of DL medical imaging solutions. Although our evaluation showed only a moderate improvement in performance metrics, we emphasize the significant potential of the methods explored to advance CS segmentation and their importance in facilitating early detection and intervention strategies for SZ and BD.
\end{abstract}


\end{frontmatter}

\section{Introduction}
\label{sec:introduction}
\subsection{Background}
Bipolar disorder (BD) and schizophrenia (SZ) are severe mental disorders that impact approximately 0.7\% and 1.0\% of the population respectively \citep{SZ_BP_prevelance}.  These conditions impose a significant burden on both individuals and society, resulting in substantial economic, mental, and societal costs \citep{SZ_burden, BP_Burden}. SZ and BD are believed to be neurodevelopmental disorders influenced by both genetic and environmental factors \citep{VIA7_Study_Protocol}. Identifying early risk markers for these diseases can enhance our understanding of their progression and lay the groundwork for primary preventive interventions.

SZ and BD typically manifest in late teenage years or early 20s, while children at familial high risk may exhibit symptoms even earlier, often before the age of 12 \citep{VIA7_Study_Protocol, SZ_BP_prevelance}. Having a family history of BD or SZ is the strongest risk factor for developing these disorders and, according to a meta-analysis, approximately 55\% of children at familial high risk will encounter mental illness in early adulthood, with around one-third experiencing severe mental illness (SMI) \citep{VIA11_Overview}.

The Danish High Risk and Resilience Study (VIA) is a longitudinal cohort study of 520 7-year-old children born to parents with schizophrenia, bipolar disorder, or no mental disorders  \citep{VIA7_Study_Protocol}. Its main objectives are to gain insights into the early disease processes of schizophrenia and bipolar disorder, investigate the developmental trajectory of children with familial high risk across various domains (neurocognition, psychopathology, social cognition, motor function) and examine the influence of genetic and environmental factors on the progression of these disorders. The study seeks to explore symptom formation, cognitive impairments, differences in brain structure and activation patterns \citep{VIA11_Overview}.
    
According to the VIA7 study results, children  born to parents diagnosed with SZ and BD already demonstrate higher rates of psychiatric diagnosis, cognitive deficits (particularly in FHR SZ), and motor difficulties by age 7 \citep{VIA7_Motor_Impariments}. When compared to controls, children with FHR of SZ show persistent developmental deficits in manual dexterity and balance.  While no observable motor development differences are found among children with FHR of BD as a group, children with definite motor problems across all groups had a higher likelihood of experiencing psychosis, suggesting a connection between childhood motor impairment and neurodevelopmental susceptibility to psychosis \citep{VIA11_Motor_Impariments}. Studying structural brain changes related to these impairments during early disease formation could provide critical information on differences in neurodevelopment between individuals with and without familial risk as well as their causes.

The central sulcus (CS) is an important landmark for examining structural brain differences in individuals with motor and sensory deficits. It is a prominent anatomical feature of the brain that separates the frontal lobe from the parietal lobe  and is symmetrically located in both hemispheres of the brain. It is one of the major sulci (grooves) found in the cerebral cortex. Research has shown that alterations in the shape and size of the central sulcus, which separates the primary motor and somatosensory areas, can impact fine motor control and sensory processing in individuals \citep{Sulci_PhD}. Therefore, analysis of the shape and morphology of the CS can contribute to a better understanding of the observed neurodevelopmental abnormalities in the FHR group.

\begin{figure}
  \begin{minipage}{\columnwidth}
    \centering
    \includegraphics[width=\linewidth]{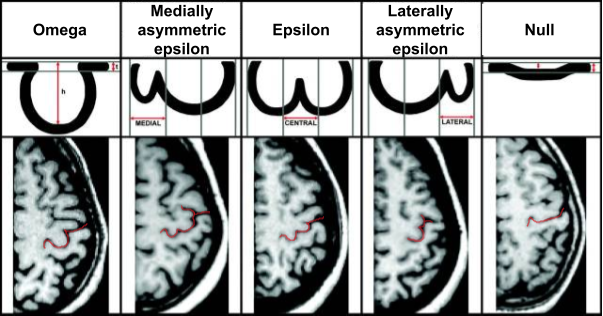}
    \caption{Schematic representation of the different morphological variants of the hand motor cortex observed in humans. Omega, medially asymmetric epsilon, laterally asymmetric epsilon, and null variants were observed in 88.3\%, 2.9\%, 7.0\%, and 1.8\% of the hemispheres, respectively with statistically significant sex differences. The epsilon variant was twice as frequent in men, and an interhemispheric concordance for morphologic variants was observed only for women. Courtesy of \citep{HandKnob_Variability}.}
    \label{fig:HandKnobVarFig}
  \end{minipage}
\end{figure}

The first step in CS analysis is its detection and segmentation, commonly based on structural magnetic resonance (MR) images. Although the central sulcus is one of the most stable and prominent folds of the human brain, its size and shape vary substantially across individuals and between hemispheres \citep{HandKnob_Variability}.  For example, one of the most prominent sections of the central sulcus is the so-called hand knob region, which has significant anatomical variations illustrated in Figure \ref{fig:HandKnobVarFig}.

Furthermore, the CS morphology depends highly on the gyrification of the cortex, which measures the degree of cortical folding \citep{GyrificationDevelopm} . Increased gyrification characterized by numerous and complex gyri and sulci may lead to more intricate and convoluted sulci patterns with more twists and turns while decreased gyrification, observed with ageing may result in shallower and less complex gyri and wider sulci \citep{GyrificationAging}. This decrease in gyrification is caused by systematic cortical thinning during normal ageing and is related to neuronal pruning, life-long reshaping and neurodegenerative processes \citep{GyrificationAging}. This in fact means that the intricate pattern of gyri and sulci will vary considerably among different age groups, particularly in children and adults as we know that the peak of gyrification happens in early childhood after which it steadily decreases over time \citep{Gyrification_progression}.

\subsection{Project proposal}
Segmenting the central sulcus poses a significant challenge due to its intrinsically high morphological variability, which is further influenced by the gyrification changes that occur with age. Successfully addressing these challenges requires sophisticated models and, crucially, large and diverse datasets that encompass the full range of CS morphological variations. Unfortunately, the only currently available dataset with manual sulci segmentations, to the best of our knowledge, is limited in terms of subject count and represents a specific cohort, making robust and precise CS segmentation on diverse populations a difficult task \citep{brainvisa}.

In light of these challenges, the primary objective of this research is to develop and investigate approaches for constructing robust and adaptable CS segmentation models.  Our experiments aim to address the issue of limited data availability and provide pipelines that can train CS segmentation models that demonstrate reliable performance on unseen and diverse populations of subjects. To achieve them, we investigate two novel ideas in the field of CS segmentation, namely how synthetic data generation can be used to model morphological variability of the CS while self-supervised pre-training and multi-task learning can be utilized for adjusting the model to new subject cohorts by learning pertinent feature representations of the cortex shape.

\section{State of the art}
\label{sec:stateoftheart}
Since the development of high-resolution brain MR imaging, the recognition of cerebral sulci and their morphology analysis has been of significant interest to researchers studying structural abnormality patterns related to the diseases affecting the neocortex \citep{EarlySInterest, EarlyInterest2}. This led to the development of several classes of approaches for automatic sulci detection.

The first type of approaches relies on feature-based elastic registration of a labelled template atlas with segmented sulci to the subject's imaging data. This method propagates labels and identifies anatomical structures of interest by matching surface features between the subject and the pre-labelled template \citep{RegSulciSegm1, RegSulciSegm2}. While these approaches have been successful in  identifying some major sulci, the high inter-subject variability of the cortical folding patterns makes it challenging to achieve an exact match between a subject and a template. Moreover, the existence of such a match is uncertain which further complicates the use of these methods for a precise sulci shape analysis. \citep{OldSulcSegmReview}.

Another set of approaches explored by \citet{CurvDepth1, CurvDepth2, CurvDepth3} consider curvature and geodesic depth properties of the cerebral folds. They use depth thresholding and deformable models to differentiate sulci and gyri using cortical surface meshes created from 3D MR images, relying on the assumption that sulci are concave and gyri are convex. However, these approaches highly depend on the ad-hoc handcrafted rules, thresholds and parameters describing the elasticity of the deformable model or depth and curvature thresholds as well as the quality of meshing, which can limit their generalizability and performance.
 
Recent advancements in image processing, computational methods, and deep learning approaches have led to substantial progress in automatic cortical sulci segmentation \citep{BrainVisaCNNPaper}. These advancements increased the accuracy of segmentation as well as expanded on the types and variety of supported sulci, enabling more precise investigations into complex folding patterns and their relationship with brain structure and function \citep{SphericalCNNPFCS}. In this section, we provide an overview of recent developments in the field, which encompass the most popular pipelines for automatic sulci segmentation and outline the motivations behind the methods explored in this study.

\subsection{Spherical CNNs}

In the past decade, deep learning models have gained significant traction in biomedical research due to their exceptional ability for feature extraction and outstanding performance \citep{DL_MedSegm}. While there have been previous efforts to apply traditional convolutional neural networks (CNNs) to segment the sulci, such as demonstrated in \citet{RegCNNParcSulc}, the unique characteristics of the convoluted cerebral cortex have led to the proposal to use a spherical variant of CNNs \citep{SphericalCNNPFCS}.

Standard 2D or 3D CNN architectures are ill-suited for handling the curved geometry of the convoluted cerebral cortex. Most CNN models are designed to optimally work in Euclidean image grids, which restricts their ability to effectively encode cortical surface data. Due to the intricate shape and high curvature of the cortex, it is possible for two points situated on the cortex to have a small Euclidean distance. However, in terms of the manifold distance through the cerebral cortex, these points could be significantly far apart representing distinct and separate regions of the brain. The complex geometry of the cortex introduces a non-linear mapping between Euclidean and manifold distances, meaning that proximity in Euclidean space does not necessarily imply proximity on the cortical surface.

These limitations have prompted the increasing popularity of spherical CNNs as they offer a more suitable framework for processing and analyzing the cortical surface \citep{SpherUsePaperEx} . However, for them to work, the cortical surface first needs to be represented as a 2D spherical manifold. This process typically involves segmenting the white matter (WM) and grey matter (GM) tissues based on structural brain images, constructing a cerebral cortex surface mesh through tesselation, and applying post-processing steps to address topological inconsistencies, holes, gaps, and optimize surface geometry \citep{FreeSuerferCortexMeshing}. Finally, the surface mesh is inflated while preserving its metric properties, resulting in an expanded, spherical representation of the cortex \citep{FS_Inflation}.

\begin{figure*}
  \centering
  \includegraphics[width=\textwidth]{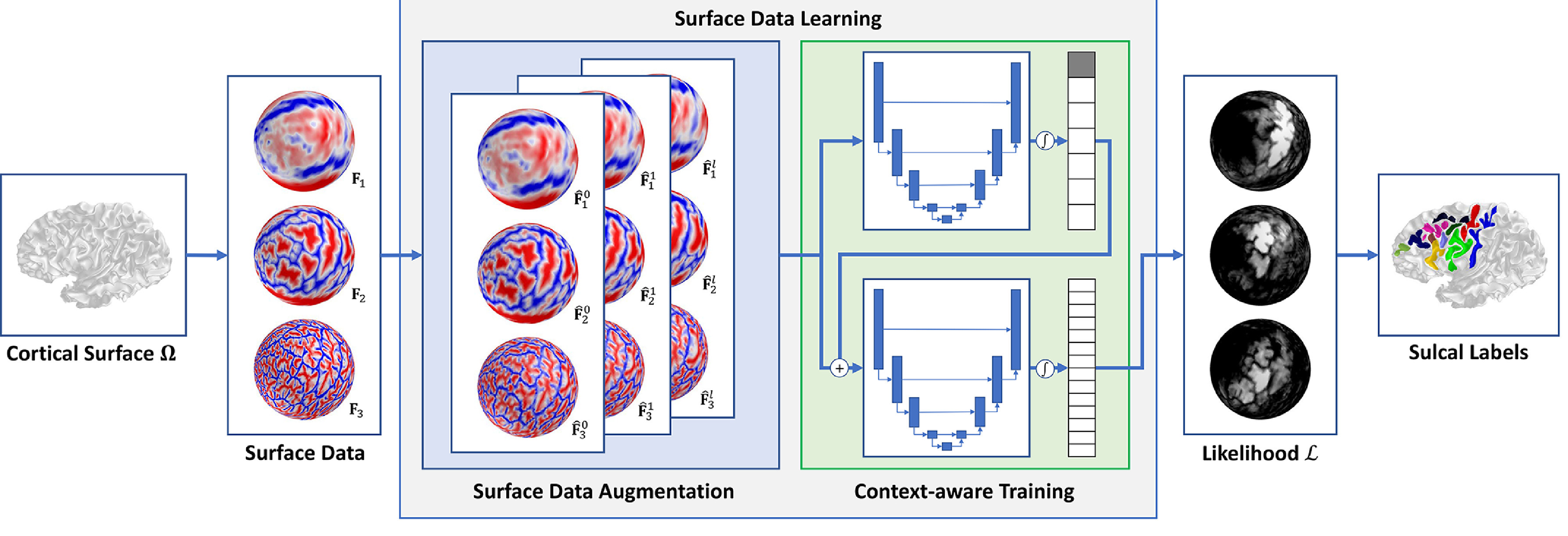}
  \caption{A schematic representation of the framework proposed by \citet{SphericalCNNPFCS}  for the training of spherical CNNs for sulci segmentation. Two main contributions are the data augmentation approach (blue box), which augments training samples by deforming them through surface registration to every possible pair of other training samples while reconstructing all intermediate deformations and using them as additional samples and the context-aware training method (green box) in which spatial information of primary/secondary sulci is extrapolated to guide the segmentation of smaller and shallower tertiary sulci. Courtesy of \citep{SphericalCNNPFCS}}
  \label{fig:spherCNNPipeline}
\end{figure*}

\citet{SphericalCNNPFCS} further improves the performance of spherical CNNs in sulci segmentation tasks by applying surface data augmentation and context-aware training in a pipeline schematically depicted in Figure \ref{fig:spherCNNPipeline}. Given the small size of the dataset used (60 and 36 in two explored cohorts), the authors emphasized a crucial need for data augmentation. However, the augmentation approaches for spherical surfaces have not been extensively explored compared to regular 2D/3D data. The authors proposed a novel approach that utilizes surface registration to augment training samples. The augmentations are achieved by applying spherical harmonics to decompose the spherical deformation needed to register every training image to all others and reconstruct intermediate deformations by controlling the basis functions. By doing so, the suggested approach bridges the gap between moving and target samples in the feature space along their deformation trajectory. This method enhances the training data by generating additional variations that improve the performance of models trained on limited samples. In their context-aware learning phase, hierarchical training is employed. The model is first trained to recognize the deeper and more stable primary sulci, and then the predicted information about their location is used as an additional input channel to guide the segmentation of shallower and more variable tertiary sulci.

While the use of spherical CNNs to capture cerebral surface topology is a promising idea, the numerous pre- and post-processing steps required to segment the tissues and generate cortical meshes and spherical surfaces present drawbacks. The performance of the separate models used in these steps can significantly impact the resulting surface representations of the cortex, leading to missed or wrongly detected sulci regions. The data augmentation technique proposed by the authors although presents a novel augmentation scheme for spherical data is nevertheless limited in its variability to sulci patterns presented in the training data. The limited amount of data augmentation techniques for spherical surfaces and the general lack of research in the field of spherical CNNs can impede the development of robust segmentation algorithms.

\subsection{Brainvisa}
\begin{figure*}
  \centering
  \includegraphics[width=\textwidth]{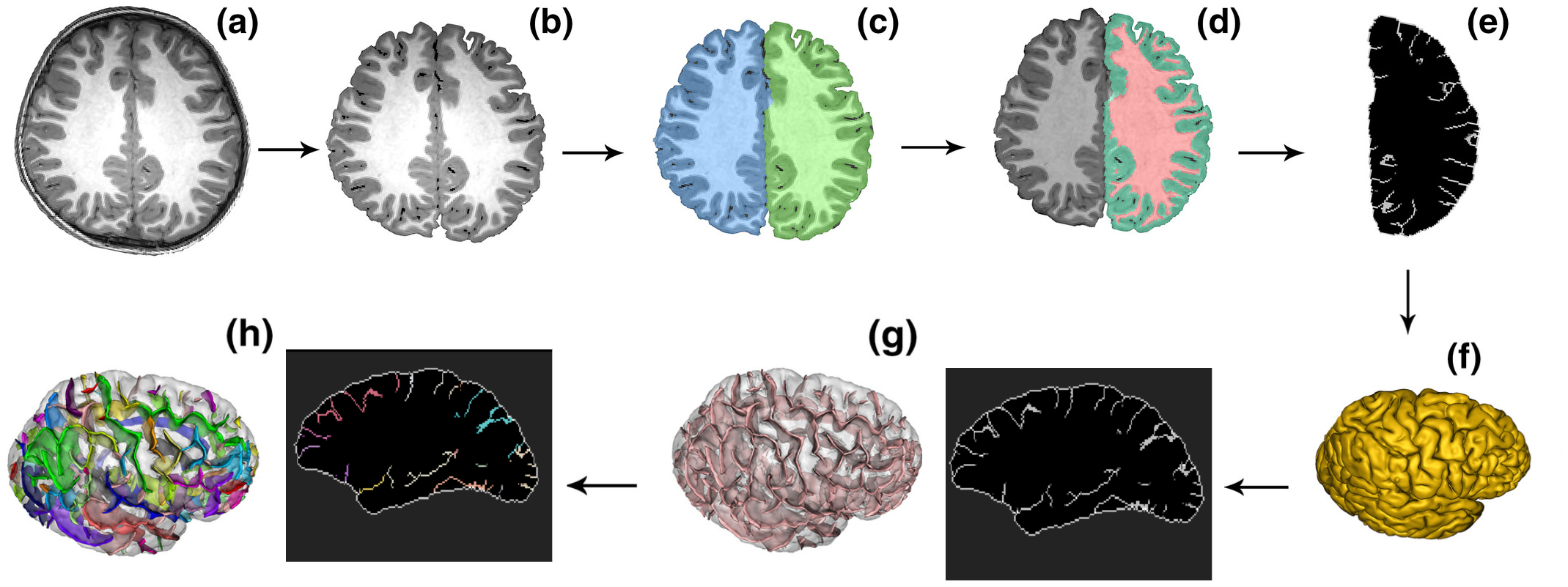}
\caption{BrainVISA pre-processing  pipeline. (a) T1w structural image; (b) Skull stripping; (c) Hemisphere segmentation; (d) GM and WM segmentation; (e) CSF skeleton labelling; (f) Cerebral cortex surface reconstruction; (g) Sulci detection; (h) Sulci parcellation. Based on \citep{BV_SS_1}.}
  \label{fig:BvisaPipeline}
\end{figure*}

The BrainVISA software package is widely recognized and utilized in the literature for sulci segmentation \citep{BVisaRef2, BVisaRef1, BVisaRef3, BV_SS_1, BvisaOldPaper,BvisaOvervewPipeline}. It offers the capability to segment more than 120 different sulci of the brain and compute morphological features based on the segmentation. In its latest version, as described in \citet{BrainVisaCNNPaper}, BrainVISA introduces several approaches for sulci labelling, consolidating decades of research in developing automatic pipelines for sulci segmentation. Although these approaches follow different directions for segmentation, they all share the same data preparation steps and begin with sulci detection. BrainVISA's pipeline encompasses multiple pre-processing steps and as shown in Figure \ref{fig:BvisaPipeline}, as it starts from a high-quality structural T1-weighted image used to first detect and then label the sulci.

\subsubsection{Pre-processing}
The pre-processing steps applied by BrainVISA aim to transform the structural MR image into a binary CSF skeleton image, where non-zero voxels define the skeleton of the CSF that corresponds to the detected sulci \citep{BrainVisaCNNPaper}. To achieve this, several key steps are carried out. The pre-processing pipeline starts with bias field correction to mitigate low-frequency intensity variations in the MRI image.  Afterwards, brain and cerebellum identification is performed, followed by the removal of non-brain tissues using a technique based on 3D erosion and template-based 3D region growth. The cortical grey matter ribbon is then obtained after which spherical meshes for the pial and GM/WM interfaces are extracted. Next, based on curvature estimation a crevasse detector reconstructs sulcal structures as medial surfaces between the two opposing gyral banks spanning from the most internal point of the sulcal fold to the cortex's convex hull. Following that, the skeleton of the CSF is fragmented into elementary folds, ensuring adherence to topological and geometric constraints specific to the sulci definition  \citep{BV_SS_1}. Finally, the CSF skeleton image is parcellated into distinct sulci using one of the following methods.

\subsubsection{Multi-atlas parcellation}

Multi-atlas segmentation (MAS) methods, originally presented by \citet{MAS}, leverage manually segmented images as atlases, wherein each atlas is adjusted to fit the image being segmented, and the best matches are selected to participate in the segmentation process. This approach enables a more accurate representation of anatomical variability by avoiding the use of an average template atlas to model the segmentation problem. Instead, MAS techniques incorporate atlases that better capture the inter-subject variability present in the data. It is worth noting, however, that the registration of atlases to the target images can be computationally demanding.

In BrainVISA, the MAS technique involves creating patches extracted as cubical slices from the training images that encompass the elementary sulci detected in the preceding steps \citep{BrainVisaCNNPaper}. These patches are then registered to the target image, where the folds skeleton has been extracted, and the best matches are determined. The patch labels are subsequently propagated onto the target image, utilizing the distance between patches to perform a robust weighted average of the labels. Finally, the propagated labels are utilized to calculate the label score maps.

\subsubsection{CNN parcellation}
Similarly to approaches based on spherical CNNs, BrainVISA's deep learning  models do not rely on the original intensity image but instead utilize a pre-processed version of it. They employ a binary 3D image that represents the skeleton of CSF. BrainVISA authors experimented with a 3D U-Net convolutional neural network (CNN) based on the architecture proposed by \cite{Cisek3D_Unet}, examining both patch-based and whole-image-based models, concluding that the U-Net model processing the entire image outperformed the patch-based one. The superiority of the whole-image approach was attributed to its ability to capture comprehensive sulcal patterns more efficiently. This was achieved by having a larger field of view and the capability to observe the complete folding pattern of the brain, enabling a better understanding of the overall structure \citep{BrainVisaCNNPaper}.

During the training process, the CSF skeleton image was used as input to the DL model, which was trained to produce a parcellation of the skeleton by assigning a specific sulci label to each non-zero voxel. During training, only the classification error of the voxels belonging to the skeleton contributed to the loss, based on the assumption that the sulci detection step was executed accurately. Such design choice reduced the complexity of the learning task, as the model solely had to learn the labelling of the skeleton voxels without considering the background. Moreover, due to the heavy reliance on pre-processed skeleton images, the researchers employed only a simple random rotation-based augmentation during training, since the binary nature of the images limited the application of complex data augmentation techniques.

\subsection{Limitations of the current methods}
While the methods utilizing spherical CNNs and BrainVISA pipelines for automated sulci segmentation demonstrate significant advancements in the field, they are not without their limitations. 

First of all, both of them have significant pre- and post-processing pipelines. The performance of individual models employed in them can substantially impact the resulting spherical surface or CSF skeleton cortex representations. Both of them heavily rely on the quality of the WM/GM/CSF segmentations that are used to build cortex meshes which could be a complicated task, especially in low-resolution images corrupted by artefacts, introducing potential inaccuracies or errors. Furthermore, in a population of children or adolescents, for example, higher cortical gyrification can lead to narrower sulci gaps which make proper differentiation between opposing gyral banks challenging due to the partial volume effect \citep{NarrowSulci}.

Secondly, both methods employ a narrow range of data augmentations, which has limited effectiveness in enhancing the diversity of cortex morphologies represented in the training set. These augmentations might fail to adequately simulate the variability of sulci that could be absent in the original data or image variation and bias induced by differences in acquisition schemes or scanners.

Finally, both approaches are trained and evaluated on small-scale in-house datasets consisting of only a few dozen images, typically representing a specific cohort. This limitation arises from the lack of comprehensive, diverse, and standardized datasets available for evaluating sulci segmentation techniques. Consequently, the generalizability and robustness of these approaches across different cohorts are called into question. 

These challenges and limitations motivate us to investigate alternative approaches in this study.  Our focus is to develope models that could effectively handle variations in image quality and contrast, which would  simplify the segmentation pipeline avoiding multiple pre-processing steps that can lead to the accumulation of errors. We are interested in exploring approaches that can efficiently utilise little available data  as well as adapt the model to diverse cohorts with previously unseen sulcal patterns. In the subsequent section, we detail the specific methodologies employed to achieve these objectives.

\section{Material and methods}
\label{sec:materialmethods}
\subsection{Datasets}
In this section, we discuss the datasets used for training and evaluation. It is important to note that the primary objective of this study is to investigate the training of robust and generalizable segmentation models. Therefore, we are exclusively using the BrainVISA dataset that has high-quality curated and labelled CS segmentations to train or fine-tune models for the CS segmentations task, while the VIA11 dataset is used solely for evaluation or self-supervised pre-training that assumes that the CS ground truths do not exist. Such a split allows us to assess the performance degradation of models trained on one dataset and evaluated on another, analysing how inherent disparities in population demographics and acquisition parameters between the datasets affect the model's performance.

To ensure uniformity in the input data for the models, we apply the same pre-processing steps for both datasets, which include only skull-stripping and registration to the common MNI template \citep{MNITemplate}. Furthermore, the images were cropped to content and resampled to a consistent resolution of 256x256x124 using the Python implementation of SimpleITK by \citet{SimpleITK}, thereby ensuring identical embedding dimensionality for VIA11 and BrainVISA images.

\subsubsection{BrainVISA}
Along with presenting the latest sulci segmentation approaches of BrainVISA, \citet{BrainVisaCNNPaper} have also released the dataset used to train them. Although it represents a significant contribution in terms of data availability, providing the first to our knowledge high-quality manually segmented dataset with multiple sulci labels for multiple subjects, it has strong limitations in terms of cohort representation. 

The dataset contains images from 62 healthy subjects selected from various databases. The subjects are predominantly right-handed men aged between 25 and 35 years. For each subject, a panel of experts produced segmentations for 63 sulci in the right hemisphere and 64 sulci in the left hemisphere through an iterative process involving consensus-based labelling, where agreement among all experts was required for the final segmentation. Although precise, such a labelling scheme excludes the possibility of  estimating inter-rater reliability. The dataset includes skull-stripped T1-weighted images, CSF skeleton images, sulci segmentations and brain masks for each subject.

The dataset was randomly split into the train (70\%) and validation (30\%) sets, enabling performance evaluation on the BrainVISA data as well. Only the training portion of the dataset was used for CS segmentation learning, synthetic data generation and self-supervised pre-training as described in the following sections.
\subsubsection{VIA11}
The VIA11 study is the second phase of the longitudinal VIA project, which focuses on assessing participants in their 11th year of life \citep{VIA11_Overview}. In contrast to the initial examination conducted at age 7 (VIA7), VIA11 study protocol incorporates several neuroimaging techniques, with our analysis focusing on structural T1-weighted (MP2RAGE) images.

For our study, we included 303 subjects who participated in the VIA11 study and had structural MR images of sufficient quality. The cohort's average age is 12.1 ± 0.28 years. It has a balanced gender distribution (49\% male, 51\% female) and includes predominantly right-handed individuals (258 right-handed, 26 left-handed, 19 ambidextrous).

Central sulcus labels for this dataset were obtained using a semi-automatic approach. First, the BrainVISA Morphologist pipeline \citep{BrainVisaCNNPaper} was employed for the initial sulci segmentation of all subjects. Then, manual quality control  was performed to estimate their correctness which resulted in 125 subjects having sufficiently good segmentations, 165 subjects having notable errors that would require manual correction of the BrainVISA segmentations, and 13 subjects having incorrect orientation or other errors that prevented manual quality control. In our work, we used 125 subjects' images for which the initial automatic tissue and sulci segmentation procedures were deemed of sufficient quality to perform self-supervised learning as well as estimate the model's performance and compare it among our approaches. The remaining 165 subjects, for which manually corrected segmentations were not available until the very end of the project were never used for any supervised or unsupervised training. These 165 images were only used as a hold-out test set for comparison between BrainVISA and our approaches. It is worth noticing nevertheless, that those manual corrections were performed based on the BrainVISA initial segmentations, and mostly consisted of removing/adding voxels to the BrainVISA's output, which in fact means that these segmentations are highly biased towards BrainVISA's output. We also note that the only pre-processing step applied to VIA11 images was skull-stripping and registration to the MNI space using the BrainVISA software, thereby replicating the same pipeline used to generate the BrainVISA dataset. This ensures uniform data representation for both the training and evaluation phases.

\subsection{Methods}

Data augmentation has emerged as a popular technique for training deep learning models in scenarios with limited training data, particularly in the medical imaging domain \citep{DataAugmentMedDL}. In this section, we explain our rationale for employing synthetic data generation based on the work by \citet{SynthSegPaper} and provide specific implementation details.

Moreover, to address the issue of limited and constrained diversity in the datasets, we investigate the use of a contrastive self-supervised framework, SimCLR, developed by \citet{SimCLRMainRef}, in conjunction with our synthetic data to learn cortex representation through self-supervised and multi-task training. We demonstrate how this approach can facilitate model adaptation to new datasets without labelled data, thereby aiding in performing segmentation tasks on dissimilar populations.

Given that our primary focus is exploring training and data generation techniques, we have opted to utilize a simple yet effective 3D CNN U-Net segmentation model designed by \citet{Cisek3D_Unet}. 3D U-Nets are among the most commonly employed architectures for 3D medical image segmentation, demonstrating effectiveness, relative computational efficiency, and robustness in the medical domain \citep{UnetMedSegmOverv}. U-Net models typically consist of symmetric encoder and decoder parts, that have skipped connections between them. The encoder part of the U-Net is responsible for extracting hierarchical and abstract feature representations from the input image, which is passed to the decoder responsible for upsampling the encoded feature maps to the original input image dimensions and generating the final segmentation map. Specifically, in all our experiments, we employed a 5-level 3D U-Net with an implementation from MONAI, \citet{cardoso2022monai}, featuring 16, 32, 64, 128, and 256 channels per layer. This choice allowed us to work with a model of comparable size and complexity to that used by \citet{BrainVisaCNNPaper}, while considering the limitations of our computational resources.

\subsubsection{Synthetic data generation}

SynthSeg, introduced in the work by \citet{SynthSegPaper}, is a segmentation model that leverages a generative approach to create synthetic images for network training. By dynamically generating training images with fully randomized parameters, the SynthSeg model learns contrast, intensity, scale, resolution, morphology, artefacts, and noise invariant features, leading to superior segmentation performance, particularly on low-quality images \citep{SynthSeg+, SynthAugm1, SynthAugm2}. We adapted the SynthSeg's data generator for our specific problem, utilizing its powerful generation capabilities to create a diverse image dataset based on the limited available labelled images.

\begin{figure*}
  \centering
  \includegraphics[width=\textwidth]{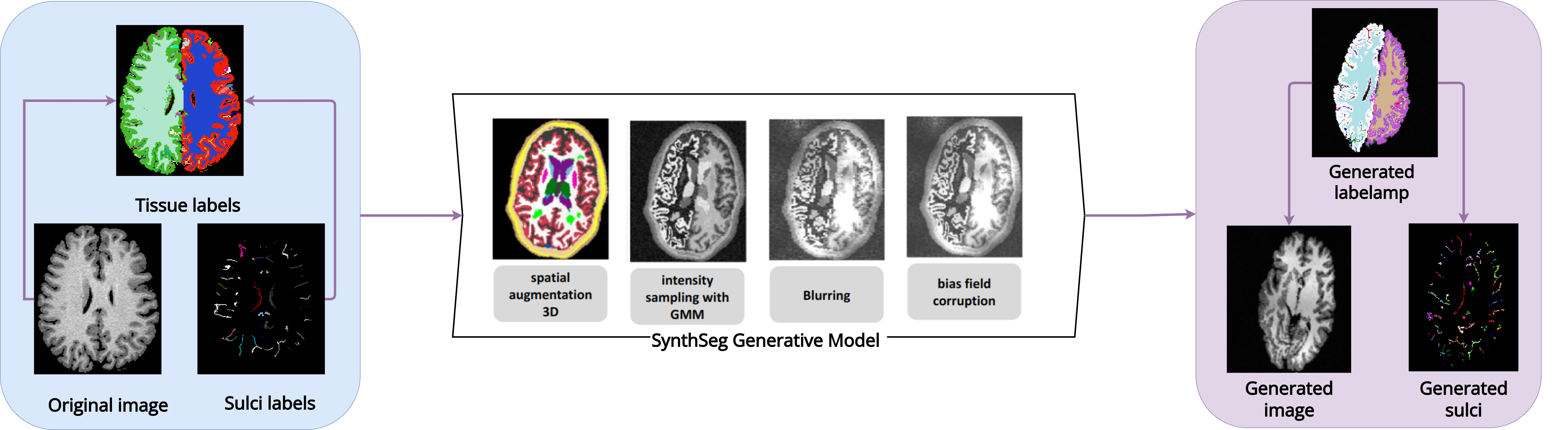}
\caption{Synthetic Data Generation Pipeline. First, we create a segmentation map, that contains both the tissue and sulci labels. Then we pass it through the SynthSeg generative model, which applies a series of transformations to the segmentation and creates the artificial image by sampling tissue-specific intensity values based on the tissue priors. Finally, the output of the model is the synthetic image and transformed segmentations that contains sulci and tissue labels.}
  \label{fig:SynthDataGen}
\end{figure*}

Figure \ref{fig:SynthDataGen} illustrates the general pipeline used to create our synthetic dataset. It starts from a segmentation (containing labels of tissues to synthesize, such as WM, GM, CSF, skull bone, and fat) that is passed as input to the SynthSeg data generator. We obtain these segmentations for our datasets from two different sources. 

For the VIA11 we utilize FreeSurfer's Samseg tool \citep{SamSeg} to obtain preliminary segmentations. These segmentations are then manually quality-checked by a skilled neuroscientist. This quality control ensures that the resulting brain segmentations are anatomically correct and could be used for subsequent image synthesis. From the 125 subjects originally reserved for SST, we select 101 that pass this quality control and only their segmentations are used for the synthetic image generation based on VIA11 data. \citet{SynthSegPaper} show that with as little as 20 segmentation maps they can reach the top performance therefore we believe that our choice of using only the 101 highest-quality segmentations will not impede the performance of the models. Generated synthetic images from the VIA11 segmentations are then used only for self-supervised pretraining described later.

For the BrainVISA dataset, we obtain the tissue segmentations by utilizing an implementation of an expectation-maximization-based algorithm described by \citet{EM_Paper}, that classifies image voxels between WM and GM based on the estimated parameters of the intensities distribution of tissues. Since BrainVISA images contain voxels belonging only to either WM or GM and skull stripping of those images is manually corrected, we opt to use this method for its simplicity and speed. For the BrainVISA segmentations, we additionally combine the central sulcus labels with the tissue labels to create a single segmentation that includes both the tissue information required for generating synthetic images and the sulci labels needed to train the CS segmentation model.

After obtaining the final segmentations we employ SynthSeg's data generator while  incorporating the following adjustments:
\begin{itemize}
\item We use T1w tissue priors provided by \citet{SynthSegPaper} to generate images with a contrast similar to T1w by sampling the intensity values for each tissue based on its Gaussian mixture model parameters. Although the original paper demonstrated that the same approach could be used to train a model invariant to any specific contrast by sampling random intensities that do not rely on any priors, we decided to use T1w-based intensities to simplify and speed up the training process, since our goal is evaluating the models' performance on the VIA11 dataset, which also consists of T1w images. Furthermore, we believe that restricting the power of possible augmentations can lead to faster convergence and the ability to learn from fewer data which is favourable in the current setting due to the limited amount of generated images and computational resources
    \item We preserve the original image dimensions when generating the output, excluding the random resampling and cropping transformations employed by the SynthSeg model. Preserving sufficient spatial resolution is crucial for accurate sulci segmentation, and reducing resolution or cropping the images may result in the loss of important information. Additionally, both the VIA11 and BrainVISA datasets contain isotropic images with a spatial resolution close to 1x1x1mm, eliminating the need to learn resolution-agnostic features for our experiments.
    \item We utilize the complete set of original spatial transformations, including random affine and elastic transformations of the segmentation map, Gaussian blurring, and bias field corruption applied to the generated image.
    \item As we use skull-stripped images, no transformations related to random drops of segmentation labels related to the skull are performed.
    \item Sulci labels are not considered during the image generation step. The model uses corresponding sulci voxels labels of background, WM, or GM for synthesizing intensities under the sulci labels, ensuring the integrity of the image and allowing potential overlap of sulci labels with WM or GM voxels, if present in the original images.
\end{itemize}

After the generation process, we obtain a pair of synthetic intensity images and the corresponding segmentation, which includes labels for the tissues and the sulci (only for BrainVISA). This dataset obtained through the generative model is referred to as the synthetic dataset.

By applying rigid and non-rigid spatial transformations, random intensity sampling, artefact generation, bias field corruption, and blurring, we simulate high variability in image appearances as well as cortex morphology while preserving crucial information necessary for CS detection and segmentation. \citet{SulcalTopology} demonstrated that consistent and accurate identification of the CS relies on several key criteria, including its relative location to other stable and distinct folds, specific shape patterns, as well as its symmetrical location and position on the cortical surface, all of which remain relatively invariant with our transformations. Therefore, by distorting the images in ways that maintain these criteria, we hypothesize that the model will learn a more robust representation of the CS location and shape, invariant to potential distortions that can occur in different datasets, caused by the changes induced by gyrification or brain volume differences,  leading to better recognition performance and increased robustness on the morphologically diverse datasets. This hypothesis is supported by the findings of \citet{SynthSeg+}, who showed the effectiveness of such synthetic data generation for tissue segmentation tasks.

\subsubsection{SimCLR}
\begin{figure}
  \begin{minipage}{\columnwidth}
    \centering
    \includegraphics[width=\linewidth]{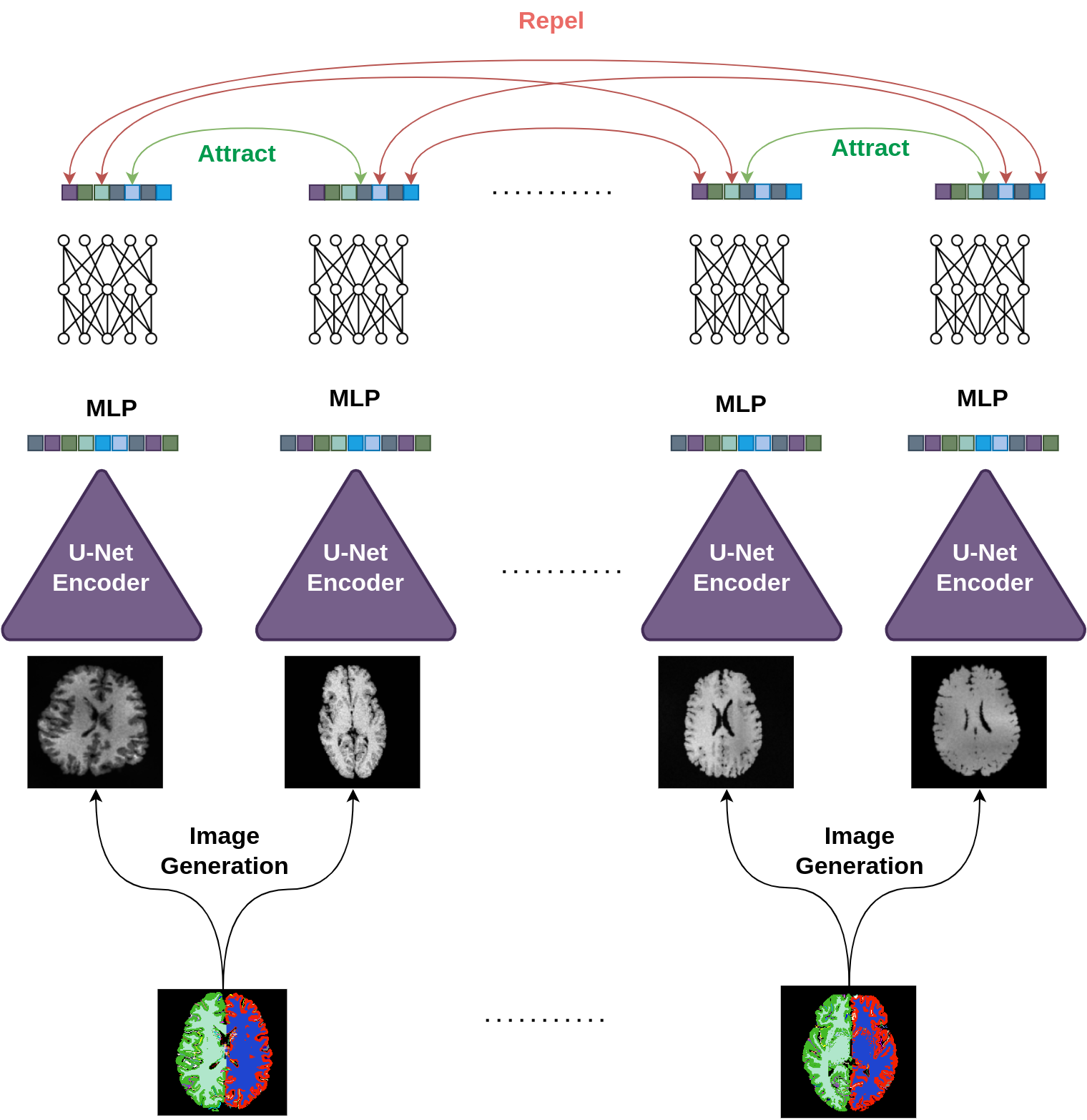}
\caption{SimCLR  framework architecture. First, two image views are generated for each segmentation present in the batch using a synthetic data generator. These synthetic images are then passed through a U-Net encoder, which calculates a dense image representation which is further projected into a space where contrastive loss is computed using an MLP. The loss function encourages the embeddings of images from the same segmentation to be close together in the embedding space while pushing apart the embeddings of images from different segmentation maps. }
    \label{fig:SSTFlow}
  \end{minipage}
\end{figure}

Self-supervised learning (SSL) is a popular method for training DL models in the absence of labelled data that has been especially popular in the medical imaging field, where the cost of labelling is extremely high \citep{SSTMedicalOverv}. Contrastive training is one of the popular SSL approaches used to learn meaningful representations of input data by maximizing the similarity between different views of the same input and minimizing the similarity between views of different data \citep{ContrSST_Revw}.  A Simple Framework for Contrastive Learning of Visual Representations (SimCLR) \citep{SimCLRMainRef} is a successful implementation of contrastive learning, particularly in medical image classification and segmentation \citep{SimCLRMedClass, SimCLRUNET1, SimCLRUNET2}. The general structure of SimCLR is illustrated in Figure \ref{fig:SSTFlow}. Our objective in utilizing SSL and SimCLR is to integrate knowledge about cortex morphology, including sulci position and shape, into the model weights during the pre-training phase. This integration is expected to be beneficial during the subsequent fine-tuning phase, where the model will focus on learning central sulcus segmentation.

The first step of the SimCLR framework involves generating multiple views of the same input image, which is a crucial step aimed at preserving relevant semantic information while introducing image variability. Random cropping, colour distortion, and Gaussian blur  proved to be effective transformations for generating views in natural image classification \citep{SimCLRMainRef}. However, in our case of 3D grayscale volumes in which cropping might erase important information about the cortex morphology we apply a different approach. Instead, we leverage the synthetic dataset generated from a single tissue segmentation and treat it as a dataset containing multiple views of the same input. We hypothesize that the diverse synthetic images derived from the same segmentation capture essential and identical information about the same cortex morphology while the unique transformations applied to each image introduce the necessary variability. This view generation process can be thought of as creating multiple distortions of the same cortical morphological fingerprint by stretching, scaling, changing its colour or elastically deforming it that would nonetheless preserve the unique pattern present in it. 

After generating the different image views, they are sequentially passed through the base model and a non-linear transformation unit based on a Multi-Layer Perceptron (MLP). The base model serves as a robust feature extractor, producing a dense representation of the image that captures key features. Inspired by recent experiments \citep{SimCLRUNET1, SimCLRUNET2}, we choose the U-Net encoder as the base model. We utilize the first five layers from the downsampling path of the U-Net and flatten the output of the last down convolution layer after max pooling to introduce it as input to the MLP. The MLP projects the feature embeddings from the base model space into a space where contrastive loss is calculated. This helps filter out specific features preferred by the contrastive loss optimization and allows the base model to learn a more robust image representation. In our experiments, we used a 3-layer MLP with a final embedding dimension of 128, as deeper MLPs have shown better results \citep{SimCLRMedClass}.

The final step involves calculating the embeddings' similarity and optimizing the total contrastive loss shown in Equation \ref{eq:SimCLRLoss}, which is based on the Normalized Temperature-scaled Cross Entropy Loss (NT-Xent) derived from the work of \citet{InfoNCE} and displayed in Equation \ref{eq:InfoNCE}. The SimCLR framework aims to maximize the similarity between the embeddings of two augmented versions of the same image (i.e., $z_i$ and $z_j$) while minimizing it between views of different images  $z_k$. The similarity between embeddings is estimated using cosine similarity defined in Equation \ref{eq:CosineSim}. 
\begin{equation}
\label{eq:SimCLRLoss}
\mathcal{L_\text{contrastive}} = \frac{1}{2N} \sum_{k=1}^{N}  [{\ell(2k-1, 2k)) + \ell({2k, 2k-1})}]
\end{equation}

\begin{equation}
\label{eq:InfoNCE}
\ell_{i,j} = -\log \frac{\exp(\text{sim}(z_i,z_j)/\tau)}{\sum_{k=1}^{2N}\mathbb{1}_{[k\neq i]}\exp(\text{sim}(z_i,z_k)/\tau)}
\end{equation}

\begin{equation}
\label{eq:CosineSim}
\text{sim}(z_i,z_j) = \frac{z_i^\top \cdot z_j}{\|z_i\| \cdot \|z_j\|}
\end{equation}

By optimizing this loss function, we aim to pre-train the model on a task similar to the downstream task, but without actual labels. The pre-training with SimCLR on synthetic data encourages the U-Net encoder to learn robust and comprehensive feature representations of the cortex morphology, as it is the only consistent and distinctive feature across different image views. Following pre-training, the model undergoes fine-tuning on the downstream task.

Our hypothesis is that by initializing the weights of the U-Net encoder with those learned during pre-training, we can transfer information about the anatomical variability of the cortical folds from a bigger and more diverse dataset and then leverage that knowledge for segmentation through fine-tuning. Specifically, we focus on pre-training with the VIA11 dataset and subsequent fine-tuning on the BrainVISA dataset to assess if the model can capture cohort-specific sulci properties that may be absent in the limited labelled data as we are especially interested in improving the performance on the VIA11 that we use for the final evaluation.

\subsubsection{Multi-task learning}
In our previous approach, we utilized the SimCLR framework for pre-training the U-Net encoder. However, we also investigate the pre-training of the decoder component in the U-Net architecture. Drawing inspiration from recent advancements in multi-task learning \citep{MultiTask1, MultiTask2}, we propose a novel pre-training framework that combines contrastive self-supervised learning with segmentation learning to pre-train the entire U-Net model.

\begin{figure*}
  \centering
  \includegraphics[width=\textwidth]{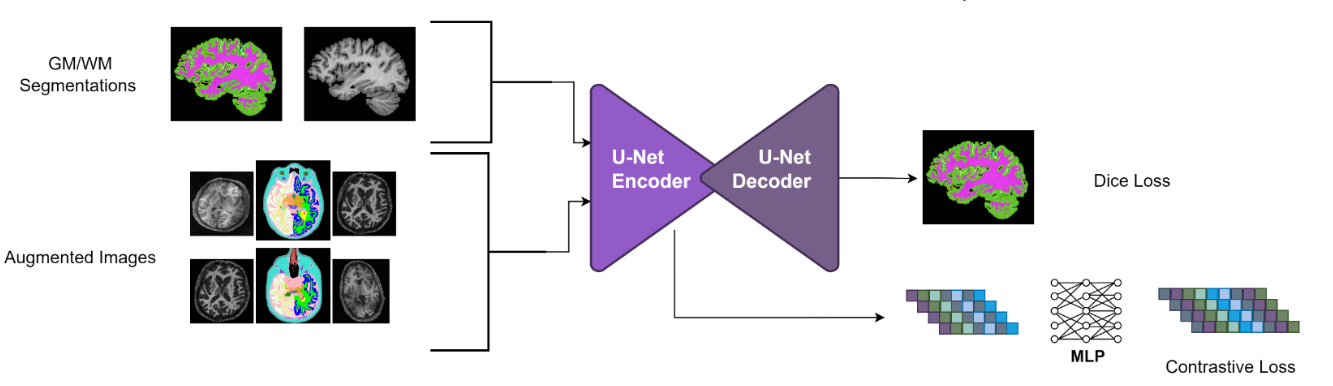}
\caption{Multi-task self-supervised training scheme. Combined contrastive and segmentation loss allows pre-training of both the encoder (on contrastive and GM segmentation tasks) and decoder (only on  GM segmentation task) of the U-Net.}
  \label{fig:MultiTaskFlow}
\end{figure*}

Illustrated in Figure \ref{fig:MultiTaskFlow}, our multi-task SSL pipeline consists of two parts. The first part follows the contrastive pre-training structure described before, calculating the contrastive loss and updating the weights of the U-Net encoder. The second part employs the same encoder model, combined with a symmetrical decoder which is simultaneously trained in a joint optimization procedure for brain tissue segmentation.  We utilize the same labels used to create synthetic images to train  the U-Net decoder to segment GM tissue based on the intensity images, effectively replicating  a part of the SynthSeg training pipeline. We choose to train the model for only single-class GM segmentation to make it compatible with the downstream single-class task of CS segmentation, avoiding the need to adjust internal embedding and kernel dimensions. Furthermore, GM segmentations were already available as a prerequisite for synthetic data generation and segmenting GM requires an understanding of the cortex morphology from the model at the surface detection level, albeit based on intensity contrast.

The final loss function optimized in this pipeline, shown in Equation \ref{eq:Multi-Task}, is a combination of the segmentation loss and the contrastive loss discussed earlier. We employ the soft dice loss implementation from MONAI \citep{cardoso2022monai} for the segmentation loss. This training scheme enables us to target the full U-Net model during the pre-training phase, learning improved weights initialization for both the encoder and decoder. Moreover, it encourages the encoder to learn representative features not only for the contrastive task but also for the segmentation task directly. By employing the multi-task loss, we aim to leverage the information learned during the pre-training phase more effectively in the downstream phase.

\begin{equation}
\label{eq:Multi-Task}
\mathcal{L_\text{multi-task}} = \mathcal{L}_{\text{segmentation}} + \mathcal{L}_{\text{contrastive}}
\end{equation}

\subsection{Training and Validation Strategy}
This section presents the technical implementation details of the tested models, training and validation strategies employed. These details aim to clarify the rationale behind our parameter choices and facilitate the replication of our results.

Unlike the approach proposed by \citet{SynthSegPaper}, which utilizes online data generation which creates synthetic images on the fly and directly feeds them into the segmentation model, we employ an offline generation approach. This choice is driven by computational limitations that prevent us from simultaneously running both the generative and segmentation models on the same GPU. Due to the storage constraints, we generate 100 synthetic images for each subject from both the BrainVISA and VIA11 datasets, resulting in 6,200 synthetic images for BrainVISA and 10,100 synthetic images for VIA11 datasets.

To ensure consistency in training parameters, we train all U-Net models for a maximum of 200 epochs during the central sulcus (CS) segmentation learning process. We employ an early stopping criterion, wherein training is halted if the validation loss fails to improve for the last 10 epochs. For CS segmentation, we adopt the Tversky Loss introduced by \citet{TverskyLoss} as our learning criterion. This loss function has demonstrated superior performance in highly imbalanced segmentation problems, which is important in our case as CS voxels occupy on average around 0.02\% of all image voxels. Although we train the model on both synthetic and original images in some experiments, validation is always performed using the original images from the corresponding validation splits. We employ a batch size of 1, as it is the maximum that can fit within our available GPUs (NVIDIA  GeForce RTX 3090 with 24GB of video RAM) and an initial learning rate (LR) of 0.001.

In the final evaluation stage,  we incorporate a post-processing step in our workflow to facilitate a meaningful comparison between the segmentations and meshes generated by our models and BrainVISA's pipeline. Given that our segmentations do not rely on CSF skeleton images and do not impose anatomical correctness requirements as part of their design, it is essential to ensure their compatibility with the meshing algorithm. To achieve this, we have opted to employ the same meshing algorithm utilized by BrainVISA in their pipeline for generating meshes from segmentations. By adopting the same tool, we minimize additional variability introduced by different meshing techniques, enabling a more accurate comparison of mesh properties.

The meshing process involves intricate calculations to create a surface based on a point cloud \citep{FreeSuerferCortexMeshing}. However, during this process, errors such as gaps, holes, and excessive tessellation can arise, particularly in the presence of noise points. To address these issues and achieve appropriate tessellation in our generated segmentations, we apply a straightforward post-processing approach. We begin by performing a morphological binary dilation on the obtained segmentation, connecting sulcus segments that are close to each other but separated spatially. Subsequently, connected component labelling is applied to the dilated image. In the final segmentation, we retain only the voxels from the original segmentation that belong to the two largest connected components calculated from the dilated image. This step ensures that only the central sulcus segments from the left and right hemispheres are retained before the meshing stage, reducing errors in the resulting sulcus mesh and enhancing its quality. We apply this post-processing step only for the final comparison between BrainVISA's pipeline and our approaches as it is needed specifically for correct meshing and fair comparison with the full BrainVISA pipeline.
 
\subsection{Quantitative Analysis}
To evaluate the quality of our segmentations, we employ two widely used metrics: the Dice similarity coefficient (DSC) and the Hausdorff distance (HD).

\begin{figure}
  \begin{minipage}{\columnwidth}
    \centering
    \includegraphics[width=\linewidth]{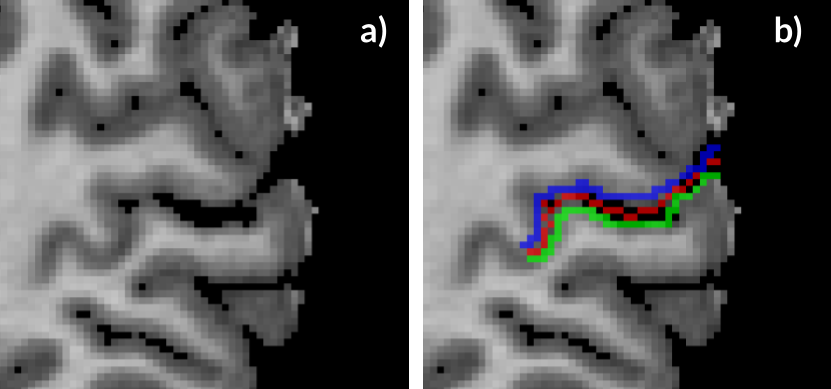}
    \caption{Ambiguity in CS segmentation. a) Brain image; b) Brain image with overlapped sulci segmentations: ground truth (red) and manually drawn alternatives (blue and green). Notice how the alternative segmentations closely follow the ground truth in terms of shape and correct anatomical position, despite having zero overlap with the ground truth and yielding a DSC of 0. The precise localization of the CS ribbon within the sulcal gap, which often spans multiple voxels in width, is inherently ambiguous. Therefore, a metric that accounts for the distance between segmentations provides a more robust measure, which makes it crucial to consider.}
    \label{fig:SegmAmbig}
  \end{minipage}
\end{figure}

The DSC quantifies the voxel-wise overlap between two segmentations, denoted as \(X\) and 
 \(Y\). Its values range from 0 to 1, where 0 represents no overlap and 1 indicates complete agreement. The DSC is computed using the following formula:

\[
DSC = \frac{{2 \times |X \cap Y|}}{{|X| + |Y|}}
\]
Another important metric we employ is the Hausdorff distance, defined as:
\[
H(X, Y) = \max \left\{ \sup_{x \in X} \inf_{y \in Y} \rho(x, y), \sup_{y \in Y} \inf_{x \in X} \rho(x, y) \right\}
\]
It measures the mutual proximity of two segmentations and provides insight into their spatial dissimilarity.  HD reflects the maximum distance of the two closest points in the segmentations and it takes positive values with smaller ones reflecting higher proximity and 0 corresponding to the complete overlap of two segmentations. Given the nature of the segmentation task, we argue that the Hausdorff distance is a crucial measure of segmentation quality that should be considered. Sulci localization is a complicated task and the precise placement of the sulci ribbon in the gap between two gyri is often ambiguous as shown in Figure \ref{fig:SegmAmbig}.

\subsection{Implementation Details}
We used Python programming language with several frameworks for this project. Pytorch and Pytorch Lightning were used to implement and train the SSL and DL models while Tensorflow was used to adapt and run the synthetic data generation pipeline. Additionally, libraries like SimpleITK-SimpleElastix and nibabel were used for image registration and spatial transformations and ITK-Snap with 3D Slicer were used for visualization purposes. 

Project code as well as other hyperparameters values and corresponding documentation can be found at: https://github.com/Vivikar/central-sulcus-analysis.

\section{Results}
In this section, we present the results of our experiments, both qualitatively and quantitatively, following the same order as in the previous section.
\label{sec:results}

\subsection{Synthetic data generation}
\begin{figure*}
  \centering
  \includegraphics[width=0.8\linewidth]{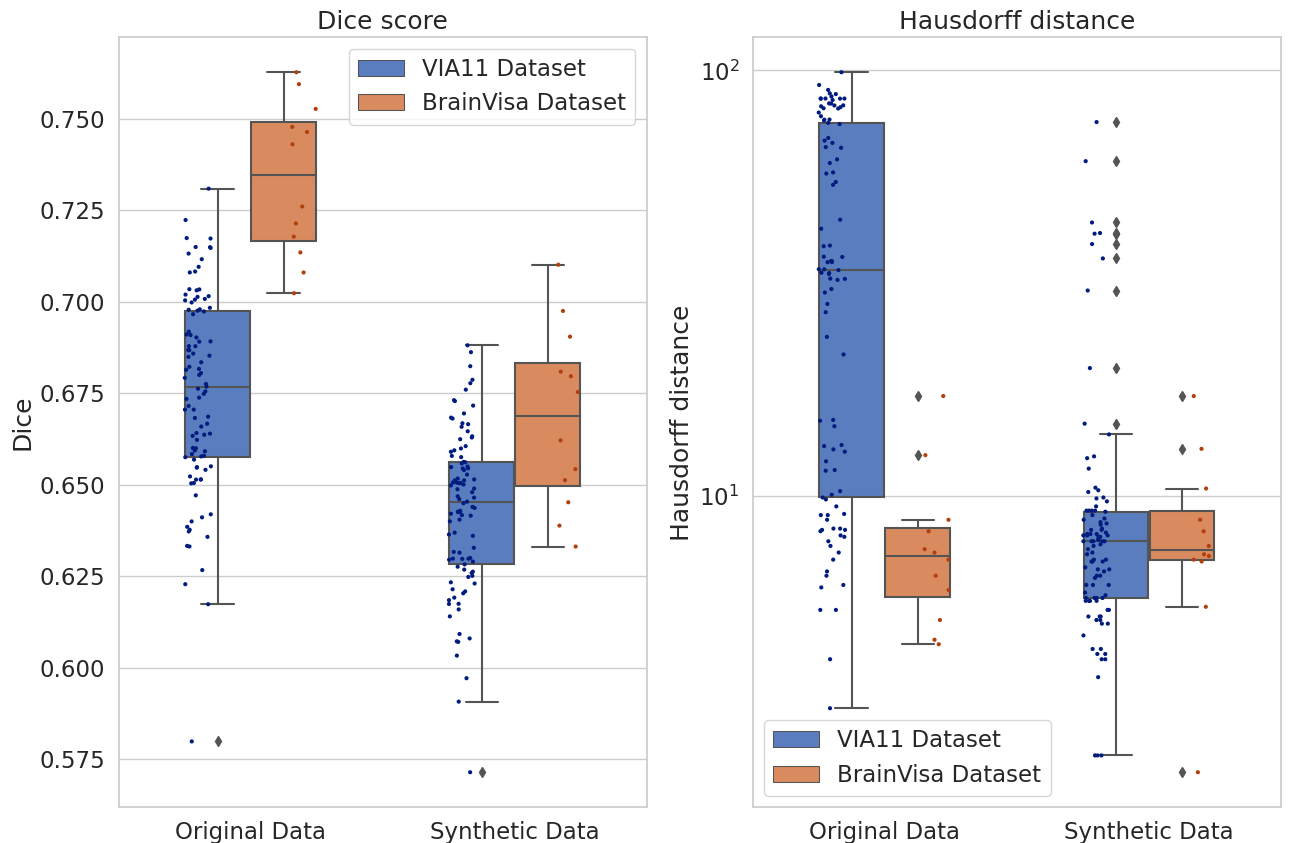}
\caption{Box plot showing DSC and HD scores for the models trained on the synthetic and original BrainVISA datasets and evaluated on the original images from the BrainVISA validation split and VIA11 dataset. A statistically significant decrease (p-value \(<0.000001\) based on a two-sided t-test) of HD scores between the model trained on synthetic and original data is observed for the VIA11 dataset.}
  \label{fig:SynthVSOrig}
\end{figure*}

We begin by examining the impact of synthetic data on the model's generalizability. To assess this, we compare the performance of the model trained on the synthetic BrainVISA dataset with the model trained on the original BrainVISA dataset. Figure \ref{fig:SynthVSOrig} displays the quantitative results comparing the performance of these two models on two evaluation datasets:  one composed of the original BrainVISA images from the validation split and the other consisting of original 125 images from the VIA11 dataset, for which we have used BrainVISA's segmentations as ground truth since they passed the quality control.

\begin{figure}
  \begin{minipage}{\columnwidth}
    \centering
    \includegraphics[width=\linewidth]{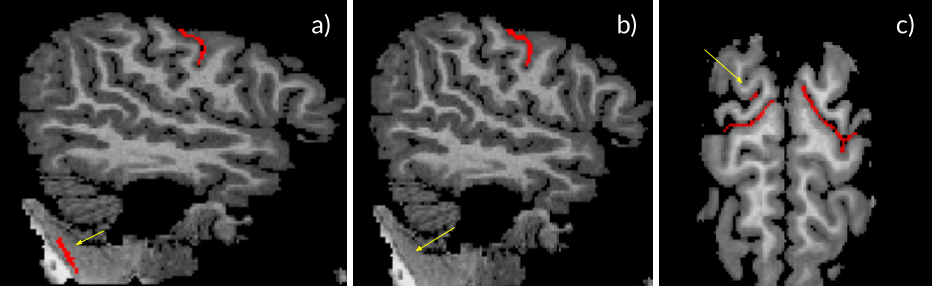}
    \caption{Sample segmentations for VIA11 subjects for the a) model trained on original BrainVISA data, and b) and c) model trained on synthetic BrainVISA data. The yellow arrows indicate the regions where some of the modes made mistakes.}
    \label{fig:Errors}
  \end{minipage}
\end{figure}

On both datasets, we observe a decrease in the Dice similarity coefficient for the models trained on synthetic data. This finding aligns with the studies conducted by \citet{SynthSegPaper} and \citet{SynthSeg+}, which demonstrate that while synthetic data yields significant improvements for images affected by artefacts, low quality, or low resolution, models trained on synthetic data tend to under-perform compared to state-of-the-art models on high-quality and high-resolution images. However, the model trained on synthetic data exhibits a substantial decrease in Hausdorff distance scores on the VIA11 dataset, which arguably provides a more sensible evaluation of performance in this setting (see Figure \ref{fig:Errors}).

Figure \ref{fig:Errors} presents qualitative results that help explain these outcomes. It illustrates how the model trained solely on the original data misclassifies the region unrelated to the central sulcus (CS), while the model trained on synthetic data does not. It is important to note that the misclassified region corresponds to the neck and represents a skull-stripping error, as it should not be present in the skull-stripped image. However,  the model trained on synthetic data makes errors in mistakenly segmenting sulci neighbouring to CS  as illustrated in image c) of Figure \ref{fig:Errors}. Although these errors lead to significantly lower Hausdorff distance scores as they are closer to ground truth, they are of great concern as they still have a substantial impact on the meshing of the CS segmentation and the subsequent estimation of its morphological features.

In our following experiments, we use the synthetic dataset for learning CS segmentation in the fine-tuning stages of the SSL as it demonstrates superior performance. 

\subsection{SimCLR}
To test how SSL can aid in adjusting the model to new datasets we apply the synthetic data generation approach described earlier to the 101 VIA11 images to create a synthetic VIA11 dataset, which we utilize for self-supervision in conjunction with the synthetic BrainVISA dataset.

\subsubsection{SSL pre-training}
\begin{figure}
  \begin{minipage}{\columnwidth}
    \centering
    \includegraphics[width=\linewidth]{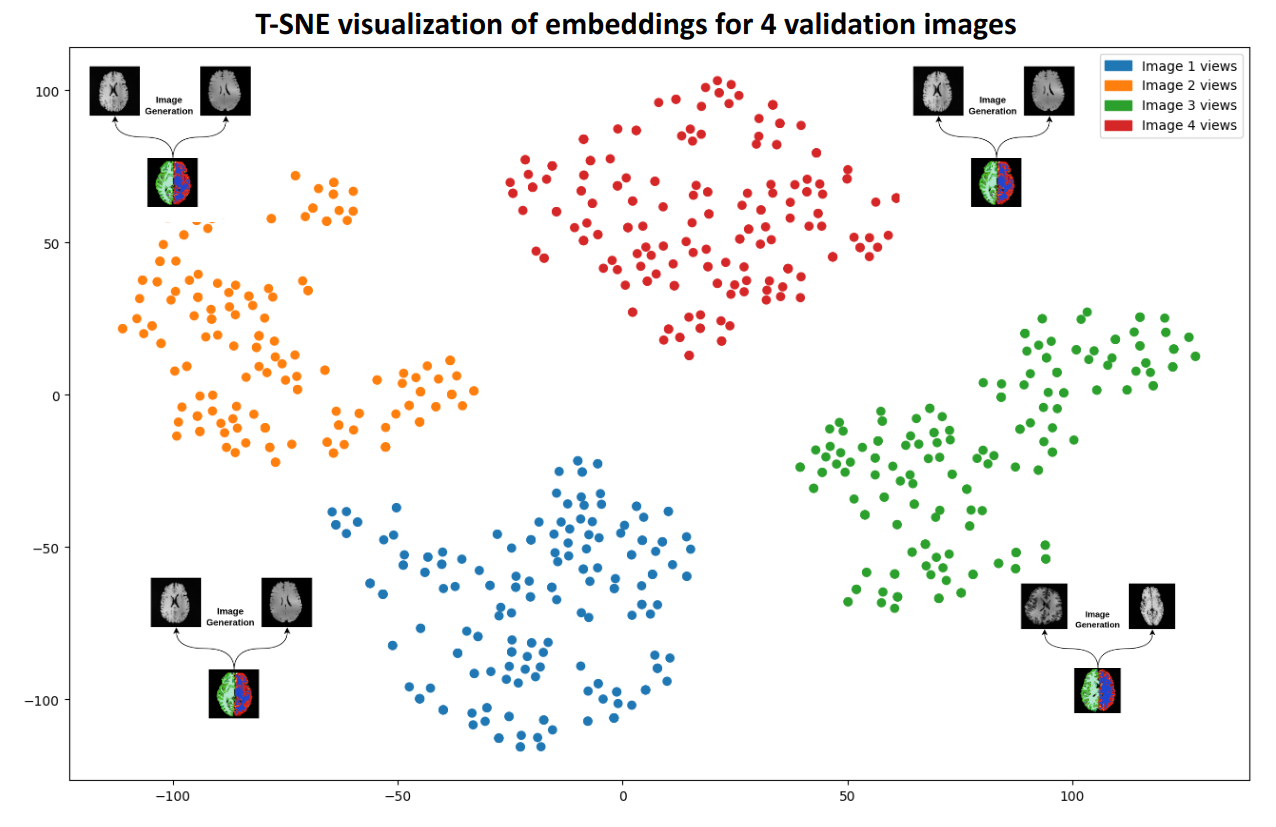}
    \caption{Visualization of embeddings of 100 different images projected onto a 2D plane with T-SNE from 4 different validation segmentations. Each coloured dot represents a synthetic image generated from a distinct segmentation.}
    \label{fig:TSNE_res}
  \end{minipage}
\end{figure}

 \citet{SimCLRMainRef} use several methods to validate the performance of their self-supervised pre-training. However, since our downstream task is not related to classification and the used images do not represent distinct categories of objects, we have chosen to validate the quality of learned image representations through the dimensionality reduction approach. Figure \ref{fig:TSNE_res} shows the projection of the embeddings outputted by the MLP from 128D space to 2D space using T-SNE \citep{TSNE}. We select random four validation VIA images that were not included in the 101 images used for self-supervised training, and we generate 100 synthetic images based on the segmentations of these four. Despite the model never encountering images generated from these four segmentations during training, we observe a clear separation between the projected embeddings corresponding to each segmentation which shows that during SSL the U-Net encoder was able to learn distinct features separating these images.

\subsubsection{Full U-Net fine-tuning}
\begin{figure*}
  \centering
  \includegraphics[width=0.8\linewidth]{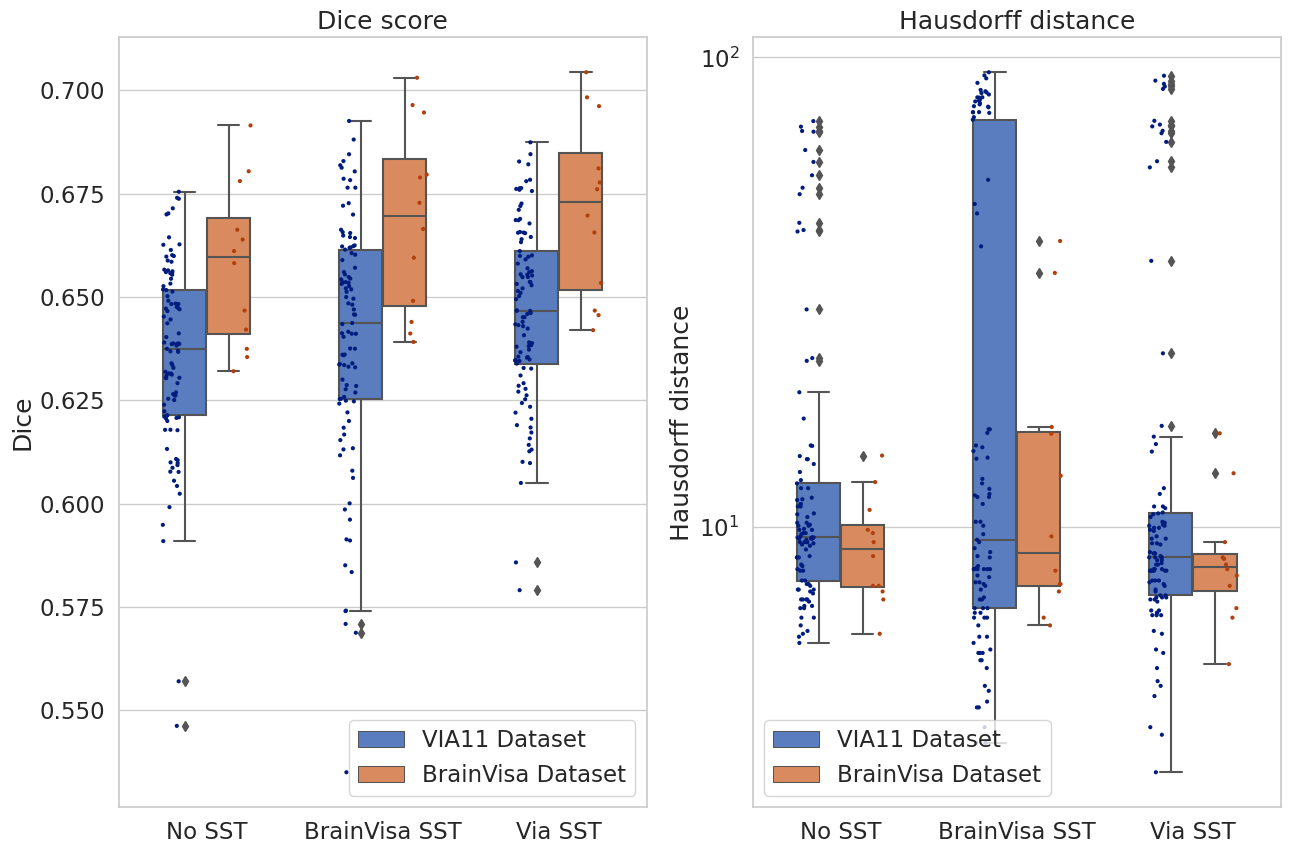}
\caption{DSC and HD for models with SSL based on different datasets with subsequent fine-tuning of the full encoder and decoder on both original datasets. A statistically significant difference (p-value \(<\) 0.005) was observed when comparing No SST with the VIA SST in terms of DSC on the VIA11 dataset.}
  \label{fig:SST_FullFineTUne}
\end{figure*}

Figure \ref{fig:SST_FullFineTUne} presents the metrics of the U-Net models with different SSL approaches, fine-tuned on the synthetic BrainVISA dataset with no frozen layers (i.e., all encoder weights were initialized with those calculated during SSL and then updated during downstream training). We observe a statistically significant increase in the Dice score metrics with SSL, with the best values for the VIA11 dataset obtained when pre-trained on VIA11 compared to no SSL (p-value=0.000799). However, we find no statistically significant difference in HD when comparing No SSL and VIA11 SSL for the VIA11 dataset. It appears that SSL with a small and homogeneous dataset like BrainVISA does not contribute to the increase of the model's generalizability and robustness, while SSL with a bigger and more diverse VIA11 dataset leads to an increase in the Dice scores without degrading HD. Therefore, in our following experiments, we consider the model pre-trained with VIA11 SSL and fine-tuned with synthetic BrainVISA data as our best-performing model.

\subsubsection{U-Net fine-tuning with the frozen encoder}
\begin{figure*}
  \centering
  \includegraphics[width=0.8\linewidth]{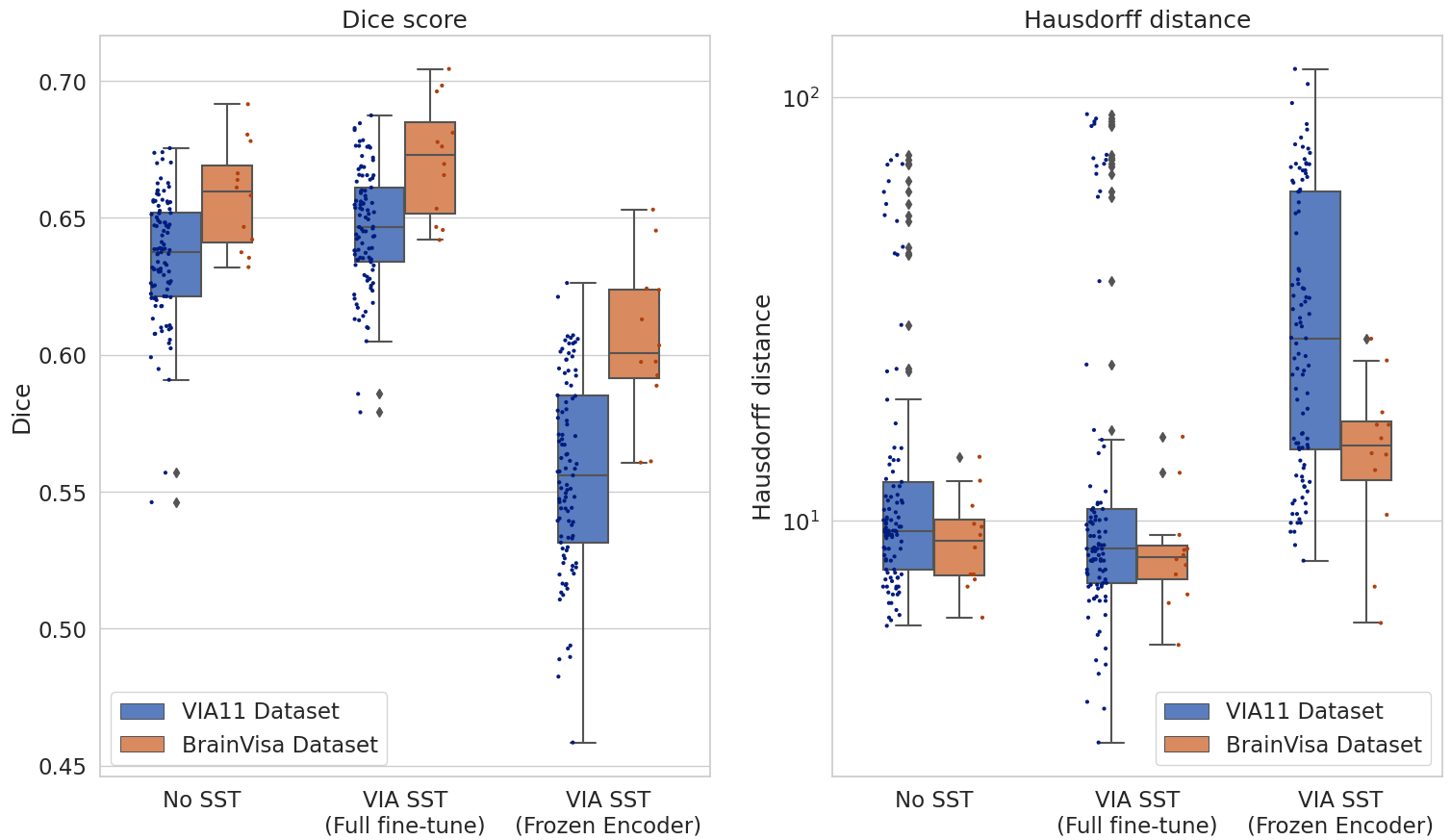}
\caption{DSC and HD for models without SSL and SSL on the VIA11 dataset with frozen and not frozen encoder on both datasets.}
  \label{fig:SST_FrozenNotComp}
\end{figure*}

\citet{SimCLRMedClass} highlights the importance of careful fine-tuning when learning downstream tasks to preserve the information learned during self-supervision. We investigate the impact of freezing the encoder model after SSL and thus completely preserving the SSL features for learning the CS segmentation task. Figure \ref{fig:SST_FrozenNotComp} compares the models with and without SSL, evaluating whether freezing the encoder during the downstream task benefits the CS segmentation performance. The results show a significant decrease in performance for both datasets and on both evaluation metrics when the encoder is frozen.

\subsection{Multi-task SSL}
\begin{figure*}
  \centering
  \includegraphics[width=0.8\linewidth]{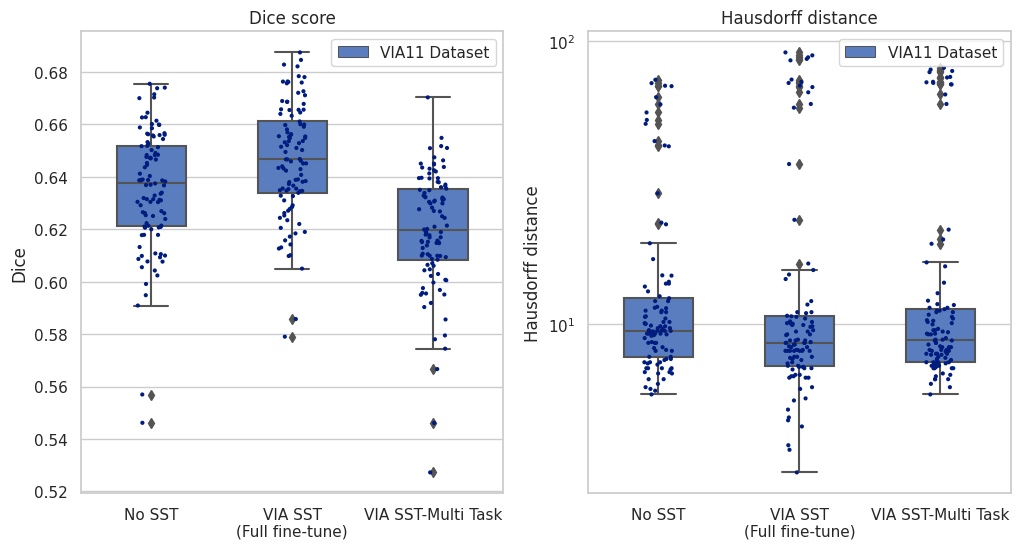}
\caption{DSC and HD for models trained without SSL, with VIA SSL and with multi-task VIA11 SSL and tested on the VIA11 dataset.}
  \label{fig:SST-MT}
\end{figure*}

As we can see in Figure \ref{fig:SST-MT} there are no statistically significant improvements in any of the metrics for the multi-task learning scenario. Due to computational limitations, we conducted experiments with SSL and evaluation solely on the VIA11 data, focusing primarily on the adaptability of our model to diverse and unseen datasets. Although no improvements were observed, we note that this strategy did not substantially degrade our results therefore such an outcome could be a result of a poor hyper-parameters selection.

\subsection{Comparison with BrainVISA}
To evaluate the effectiveness of our approach and compare it with the state-of-the-art BrainVISA's pipeline, we conducted several experiments using 165 VIA11 images from the held-out test set. These images were not utilized during any stage of the pre-training or data synthesis. We have obtained manual ground truth CS segmentations by correcting the initial BrainVISA's pipeline output for them, as it was deemed necessary during our initial quality assessment of the BrainVISA's results.

\begin{figure*}
  \centering
  \includegraphics[width=0.7\linewidth]{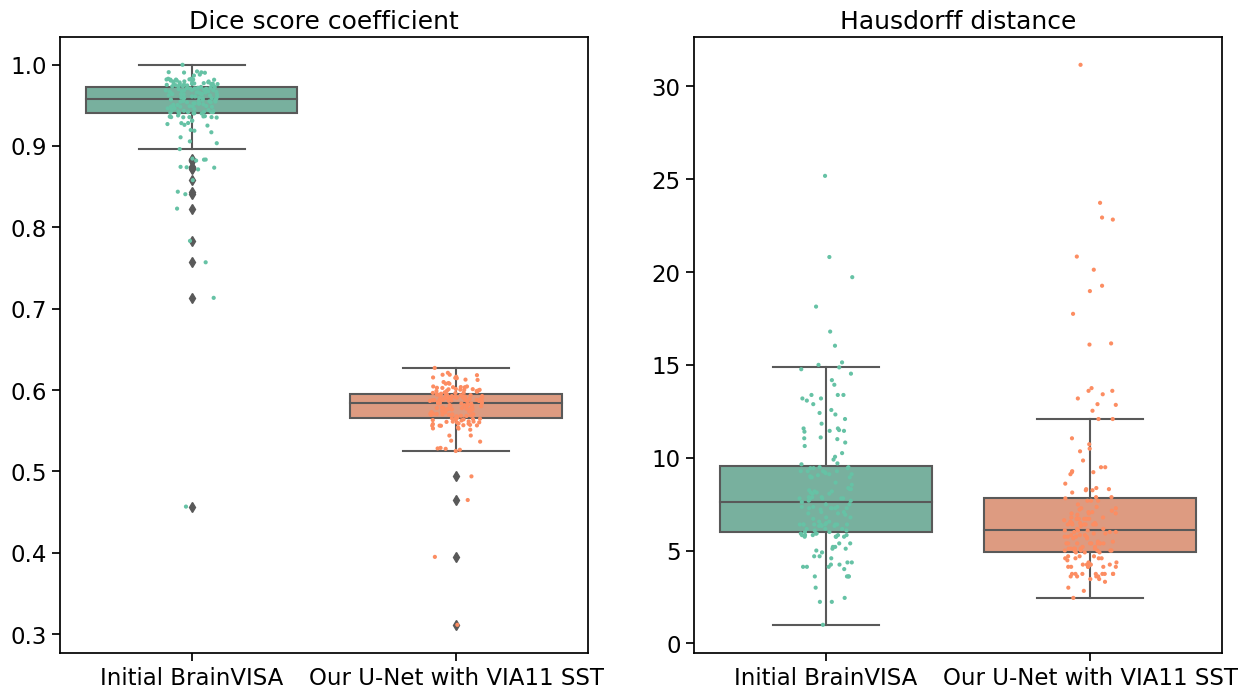}
  \caption{DSC and HD scores of the BrainVISA segmentations and our U-Net model trained with VIA11 SSL. Our model shows a statistically significant decrease in the HD scores with a p-value of 0.0379 based on a two-sample T-test.}
  \label{fig:BvisaSSTComp}
\end{figure*}

Figure \ref{fig:BvisaSSTComp} presents a comparison between our best model that is based on VIA11 SST with the further fine-tuning on the synthetic BrainVISA data. We see that BrainVISA's segmentations have a much higher Dice score.

\begin{figure*}
  \centering
  \includegraphics[width=0.8\linewidth]{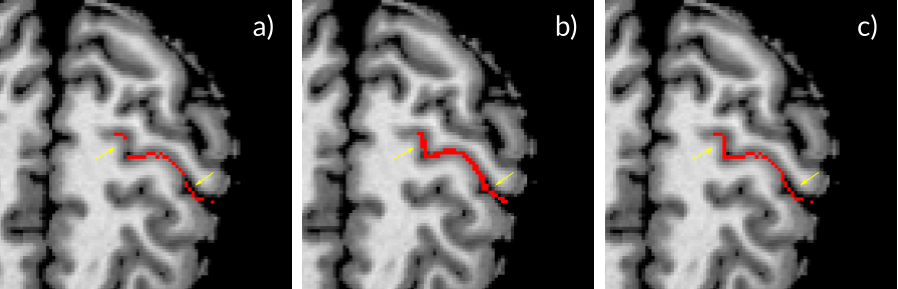}
\caption{Segmentation examples.  a) produced by BrainVISA software, b) produced by our VIA11 SST U-Net model, c) manually corrected ground truth. Yellow arrows indicate gaps in the segmentation ribbon that can lead to holes in the resulting mesh interfering with morphological features calculation.}
  \label{fig:SegmQualComp}
\end{figure*}

This substantial difference in DSC can be attributed to two main factors. First, the manual segmentations are essentially modifications of BrainVISA results, and in many cases, the initial BrainVISA estimate, if sufficiently accurate, was left unchanged. Second, the nature of the segmentations generated by our algorithm is characterized by thicker segmentation ribbons as can be seen in Figure  \ref{fig:SegmQualComp}, which are anatomically and morphologically correct but receive lower Dice scores due to its voxel-wise intersection evaluation. However, our approach shows a statistically significant improvement in the HD (BrainVISA's mean  8.315 vs our model's 7.37 with p-value \(<\) 0.005). 

\begin{figure*}
  \centering
  \includegraphics[width=0.6\linewidth]{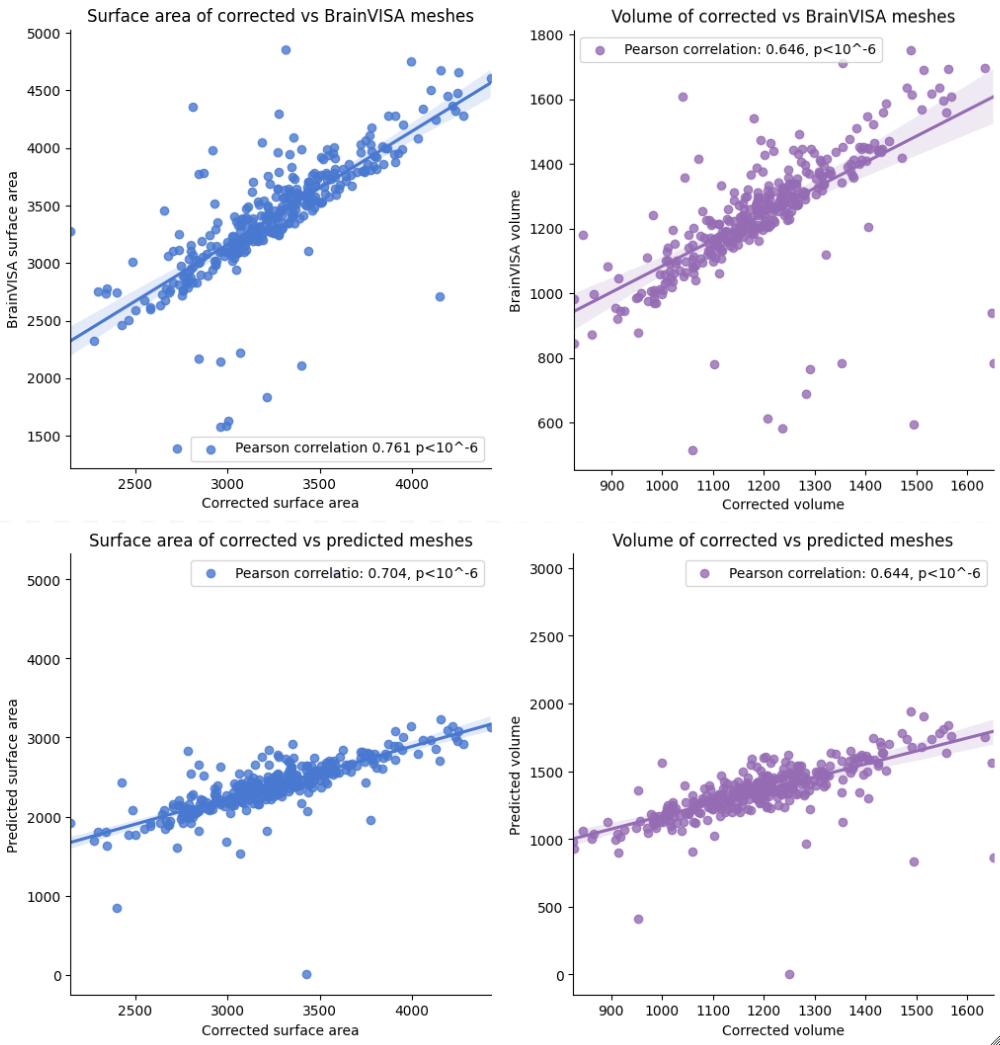}
\caption{Volume and surface area of the meshes calculated based on the manually corrected segmentation, BrainVISA's and ours (predicted by the VIA11 SST U-Net) plotted against each other.}
  \label{fig:VolAreaCorr}
\end{figure*}

Considering our ultimate goal of evaluating morphological features of the CS, an essential step in their analysis is meshing and subsequent extraction of shape features. The BrainVISA software provides built-in tools for meshing sulci segmentations. We utilized these tools to compute meshes for the segmentations obtained from the BrainVISA pipeline, manually corrected segmentations, and segmentations produced by our best model. Figure \ref{fig:VolAreaCorr} displays correlation plots between the volume and surface area calculated from these three meshes. It is immediately apparent that the morphological features of volume and surface area calculated from our segmentations exhibit a close correlation with those calculated from the manual segmentations.

\section{Discussion and conclusions}
\label{sec:discussion}
In this study, we have presented and evaluated various approaches for training deep learning models to perform central sulcus segmentation. Our review of the current state-of-the-art approaches revealed important limitations in models trained on small and restricted labelled datasets, which fail to account for the neuroanatomical variability of cortical morphology. We have emphasized the need for robust and automatic segmentation models and proposed novel frameworks to address this challenge. Our frameworks focus on two key ideas: efficient utilization of limited labelled data through artificial simulation of cortical variability in synthetic images and the creation of a pipeline for adapting the model to new subject populations through self-supervised learning of cortex morphology features.

\subsection{Discussion of results}

Firstly, we explored the use of synthetic data to train more robust models. Our findings indicate that models trained with synthetic data exhibit significantly lower Hausdorff distance (HD) scores on the VIA11 dataset, despite never being exposed to it during training. This dataset consists of a different subject population, with different image contrasts, and quality compared to the training dataset. This promising result highlights the potential of using synthetic data to simulate the morphological variability of cortex present in diverse subject populations that is beneficial for sulci localization. This approach helps address the limitations posed by small datasets, which have historically hindered progress in sulci segmentation research.

We tested the effectiveness of a self-supervised learning framework based on SimCLR combined with synthetic data for learning unique and distinct representations of cortex morphology. Our experiments demonstrate that our pre-training strategy for the U-Net encoder leads to improved DSC for the VIA11 dataset. This shows that with this pipeline we can adapt our segmentation model to new datasets without requiring any labels for them, effectively transferring information about cortex shape variability from the new dataset to our model. Consequently, this approach shows that we can improve the segmentation results for new cohorts by performing SSL on just the intensity images, paving the way to utilizing abundant unlabelled datasets that are openly available for the training of foundation models that better capture real-world anatomical variability of the cortex.

Although the multi-task SSL approach did not substantially improve our results, we believe that more careful pre-training of both the encoder and decoder models could yield better performance. We did not extensively experiment with hyper-parameter tuning for the multi-task framework and subsequent fine-tuning, which could explain the lack of improvement. Additionally, the substantial difference between GM segmentation and CS segmentation may have hindered the adaptability of the SSL loss for CS segmentation.

Lastly, we compared our best model with the current state-of-the-art pipeline from BrainVISA and demonstrated comparable performance with an improvement in the HD metric. To validate the correctness of the morphological structures represented by these segmentations, we constructed meshes based on them. The meshes built from segmentations produced by our models have highly correlated surface area and volume measures to those obtained from manual ground truths, indicating that the developed approach can be effectively utilized for CS segmentation that can be further used for analyzing the shape properties of the central sulcus.

\subsubsection{Limitations and future work}
The objective of this study was to explore and establish a proof of concept for training robust CS segmentation models directly from intensity images, without the need for extensive pre- or post-processing steps. We aimed to address the challenges commonly encountered in the medical imaging domain, including the limited availability of labelled data and the high morphological variability of the target structure.

However, it is important to acknowledge the limitations of our work. We conducted evaluations on only one external dataset (VIA11) in addition to the training dataset (BrainVISA). To obtain a more comprehensive understanding of the pipeline's performance and robustness, further evaluations on additional datasets with diverse population cohorts are necessary. For instance, testing the model on datasets consisting of elderly individuals with neurodegenerative processes resulting in substantial brain atrophy or infants and young children with either under-developed or over-developed cortical gyrification could provide valuable insights.

In our synthetic data generation process, we utilized a limited set of artificial images due to storage and computational constraints. To enhance the diversity of the synthetic dataset, implementing an online generation procedure with unlimited and unique images for each generation could be explored. Additionally, we were unable to extensively experiment and determine an optimal set of transformation parameters for the SynthSeg Generative model. We believe that exhaustive tuning of spatial and intensity parameters applied to the images can further increase the dataset's diversity. Moreover, incorporating brain images without skull stripping could improve the model's robustness to potential artefacts that can occur if it is performed with errors as well as potentially eliminate the need for this pre-processing step.

In the SSL training stage, \citet{SimCLRMainRef} demonstrated that large batch size leads to better performance. Although  longer training can partially mitigate the effects of smaller batch sizes, we did not specifically study how batch size affects our SSL stage. Moreover, in the multi-task SSL setting, we did not explore weighting schemes for the contrastive and segmentation losses due to time and GPU constraints, despite studies suggesting potential benefits \citep{WeightedMT}.

Finally, we have not experimented with different DL architectures for our base segmentation model, although there are many new architectures based on Transformers that show promising results in the medical image segmentation field \citep{DL_MedSegm} as well as U-Net variations. We have chosen to use a simple and lightweight U-Net model that made possible an extensive exploration of the proposed solutions given our computational constraints.

\subsection{Conclusions}
Synthetic data generation and self-supervised learning are two powerful tools that can address challenges in the development and deployment of DL models for recognition and segmentation tasks. In this study, we have demonstrated that by employing synthetic data within a self-supervised learning framework that enables the model to learn unique cortical morphology representations, we can achieve results that are comparable to state-of-the-art methods in central sulcus segmentation. These approaches alleviate the need for costly and error-prone pre-processing steps, allowing the training of robust and generalizable DL models that can be adapted to new cohorts without requiring any ground truth labels and work efficiently even with little available training data.

\section*{Acknowledgements}

First and foremost, I would like to express gratitude to my supervisor, Kristoffer Madsen, for his guidance and invaluable support throughout this project. I would also like to express my thanks to my colleagues at DRCMR, Enedino Hernández-Torres and Line K. Johnsen, for their assistance with data access, quality estimations, and manual segmentations of the sulci data. I am also grateful to Melissa Larsen and the centre's director Hartwig R. Siebner, for helping me better understand the neurobiological basis of this project as well as for providing me with a great opportunity to work at DRCMR.

\bibliography{myThesisBibfile}

\begin{thebibliography}{61}
\expandafter\ifx\csname natexlab\endcsname\relax\def\natexlab#1{#1}\fi
\providecommand{\url}[1]{\texttt{#1}}
\providecommand{\href}[2]{#2}
\providecommand{\path}[1]{#1}
\providecommand{\DOIprefix}{doi:}
\providecommand{\ArXivprefix}{arXiv:}
\providecommand{\URLprefix}{URL: }
\providecommand{\Pubmedprefix}{pmid:}
\providecommand{\doi}[1]{\href{http://dx.doi.org/#1}{\path{#1}}}
\providecommand{\Pubmed}[1]{\href{pmid:#1}{\path{#1}}}
\providecommand{\bibinfo}[2]{#2}
\ifx\xfnm\relax \def\xfnm[#1]{\unskip,\space#1}\fi
\bibitem[{Azizi et~al.(2021)Azizi, Mustafa, Ryan, Beaver, Freyberg, Deaton,
  Loh, Karthikesalingam, Kornblith, Chen et~al.}]{SimCLRMedClass}
\bibinfo{author}{Azizi, S.}, \bibinfo{author}{Mustafa, B.},
  \bibinfo{author}{Ryan, F.}, \bibinfo{author}{Beaver, Z.},
  \bibinfo{author}{Freyberg, J.}, \bibinfo{author}{Deaton, J.},
  \bibinfo{author}{Loh, A.}, \bibinfo{author}{Karthikesalingam, A.},
  \bibinfo{author}{Kornblith, S.}, \bibinfo{author}{Chen, T.}, et~al.,
  \bibinfo{year}{2021}.
\newblock \bibinfo{title}{Big self-supervised models advance medical image
  classification}, in: \bibinfo{booktitle}{Proceedings of the IEEE/CVF
  International Conference on Computer Vision}, pp.
  \bibinfo{pages}{3478--3488}.
\bibitem[{Behnke et~al.(2003)Behnke, Rettmann, Pham, Shen, Resnick, Davatzikos
  and Prince}]{RegSulciSegm2}
\bibinfo{author}{Behnke, K.J.}, \bibinfo{author}{Rettmann, M.E.},
  \bibinfo{author}{Pham, D.L.}, \bibinfo{author}{Shen, D.},
  \bibinfo{author}{Resnick, S.M.}, \bibinfo{author}{Davatzikos, C.},
  \bibinfo{author}{Prince, J.L.}, \bibinfo{year}{2003}.
\newblock \bibinfo{title}{Automatic classification of sulcal regions of the
  human brain cortex using pattern recognition}, in:
  \bibinfo{booktitle}{Medical imaging 2003: Image processing},
  \bibinfo{organization}{SPIE}. pp. \bibinfo{pages}{1499--1510}.
\bibitem[{Billot et~al.(2020)Billot, Greve, Van~Leemput, Fischl, Iglesias and
  Dalca}]{SynthAugm1}
\bibinfo{author}{Billot, B.}, \bibinfo{author}{Greve, D.},
  \bibinfo{author}{Van~Leemput, K.}, \bibinfo{author}{Fischl, B.},
  \bibinfo{author}{Iglesias, J.E.}, \bibinfo{author}{Dalca, A.V.},
  \bibinfo{year}{2020}.
\newblock \bibinfo{title}{A learning strategy for contrast-agnostic mri
  segmentation}.
\newblock \bibinfo{journal}{arXiv preprint arXiv:2003.01995} .
\bibitem[{Billot et~al.(2023a)Billot, Greve, Puonti, Thielscher, Van~Leemput,
  Fischl, Dalca and Iglesias}]{SynthSegPaper}
\bibinfo{author}{Billot, B.}, \bibinfo{author}{Greve, D.N.},
  \bibinfo{author}{Puonti, O.}, \bibinfo{author}{Thielscher, A.},
  \bibinfo{author}{Van~Leemput, K.}, \bibinfo{author}{Fischl, B.},
  \bibinfo{author}{Dalca, A.V.}, \bibinfo{author}{Iglesias, J.E.},
  \bibinfo{year}{2023}a.
\newblock \bibinfo{title}{Synthseg: {Segmentation} of brain {MRI} scans of any
  contrast and resolution without retraining}.
\newblock \bibinfo{journal}{{Medical} {Image} {Analysis}} \bibinfo{volume}{86},
  \bibinfo{pages}{102789}.
\newblock \DOIprefix\doi{10.1016/j.media.2023.102789}.
\bibitem[{Billot et~al.(2023b)Billot, Magdamo, Cheng, Arnold, Das and
  Iglesias}]{SynthSeg+}
\bibinfo{author}{Billot, B.}, \bibinfo{author}{Magdamo, C.},
  \bibinfo{author}{Cheng, Y.}, \bibinfo{author}{Arnold, S.E.},
  \bibinfo{author}{Das, S.}, \bibinfo{author}{Iglesias, J.E.},
  \bibinfo{year}{2023}b.
\newblock \bibinfo{title}{Robust machine learning segmentation for large-scale
  analysis of heterogeneous clinical brain mri datasets}.
\newblock \bibinfo{journal}{Proceedings of the National Academy of Sciences}
  \bibinfo{volume}{120}, \bibinfo{pages}{e2216399120}.
\newblock \DOIprefix\doi{10.1073/pnas.2216399120}.
\bibitem[{Borne et~al.(2020)Borne, Rivi{\`{e}}re, Mancip and
  Mangin}]{BrainVisaCNNPaper}
\bibinfo{author}{Borne, L.}, \bibinfo{author}{Rivi{\`{e}}re, D.},
  \bibinfo{author}{Mancip, M.}, \bibinfo{author}{Mangin, J.F.},
  \bibinfo{year}{2020}.
\newblock \bibinfo{title}{Automatic labeling of cortical sulci using patch- or
  {CNN}-based segmentation techniques combined with bottom-up geometric
  constraints}.
\newblock \bibinfo{journal}{Medical Image Analysis} \bibinfo{volume}{62},
  \bibinfo{pages}{101651}.
\newblock \DOIprefix\doi{10.1016/j.media.2020.101651}.
\bibitem[{Brainvisa(2019)}]{brainvisa}
\bibinfo{author}{Brainvisa}, \bibinfo{year}{2019}.
\newblock \bibinfo{title}{Sulci database}.
\newblock \bibinfo{howpublished}{Online}.
\bibitem[{Burton et~al.(2023)Burton, Krantz, Skovgaard, Brandt, Gregersen,
  S{\o}ndergaard, Knudsen, Andreassen, Veddum, Rohd, Wilms, Tjott, Hjorth{\o}j,
  Ohland, Greve, Hemager, Bliksted, Mors, Plessen, Thorup and
  Nordentoft}]{VIA11_Motor_Impariments}
\bibinfo{author}{Burton, B.K.}, \bibinfo{author}{Krantz, M.F.},
  \bibinfo{author}{Skovgaard, L.T.}, \bibinfo{author}{Brandt, J.M.},
  \bibinfo{author}{Gregersen, M.}, \bibinfo{author}{S{\o}ndergaard, A.},
  \bibinfo{author}{Knudsen, C.B.}, \bibinfo{author}{Andreassen, A.K.},
  \bibinfo{author}{Veddum, L.}, \bibinfo{author}{Rohd, S.B.},
  \bibinfo{author}{Wilms, M.}, \bibinfo{author}{Tjott, C.},
  \bibinfo{author}{Hjorth{\o}j, C.}, \bibinfo{author}{Ohland, J.},
  \bibinfo{author}{Greve, A.}, \bibinfo{author}{Hemager, N.},
  \bibinfo{author}{Bliksted, V.F.}, \bibinfo{author}{Mors, O.},
  \bibinfo{author}{Plessen, K.J.}, \bibinfo{author}{Thorup, A.A.E.},
  \bibinfo{author}{Nordentoft, M.}, \bibinfo{year}{2023}.
\newblock \bibinfo{title}{Impaired motor development in children with familial
  high risk of schizophrenia or bipolar disorder and the association with
  psychotic experiences: a 4-year danish observational follow-up study}.
\newblock \bibinfo{journal}{The Lancet Psychiatry} \bibinfo{volume}{10},
  \bibinfo{pages}{108--118}.
\newblock \DOIprefix\doi{10.1016/s2215-0366(22)00402-3}.
\bibitem[{Burton et~al.(2017)Burton, Thorup, Jepsen, Poulsen, Ellersgaard,
  Spang, Christiani, Hemager, Gantriis, Greve, Mors, Nordentoft and
  Plessen}]{VIA7_Motor_Impariments}
\bibinfo{author}{Burton, B.K.}, \bibinfo{author}{Thorup, A.A.E.},
  \bibinfo{author}{Jepsen, J.R.}, \bibinfo{author}{Poulsen, G.},
  \bibinfo{author}{Ellersgaard, D.}, \bibinfo{author}{Spang, K.S.},
  \bibinfo{author}{Christiani, C.J.}, \bibinfo{author}{Hemager, N.},
  \bibinfo{author}{Gantriis, D.}, \bibinfo{author}{Greve, A.},
  \bibinfo{author}{Mors, O.}, \bibinfo{author}{Nordentoft, M.},
  \bibinfo{author}{Plessen, K.J.}, \bibinfo{year}{2017}.
\newblock \bibinfo{title}{Impairments of motor function among children with a
  familial risk of schizophrenia or bipolar disorder at 7 years old in denmark:
  an observational cohort study}.
\newblock \bibinfo{journal}{The Lancet Psychiatry} \bibinfo{volume}{4},
  \bibinfo{pages}{400--408}.
\newblock \DOIprefix\doi{10.1016/s2215-0366(17)30103-7}.
\bibitem[{Cardoso et~al.(2022)Cardoso, Li, Brown, Ma, Kerfoot, Wang, Murrey,
  Myronenko, Zhao, Yang, Nath, He, Xu, Hatamizadeh, Myronenko, Zhu, Liu, Zheng,
  Tang, Yang, Zephyr, Hashemian, Alle, Darestani, Budd, Modat, Vercauteren,
  Wang, Li, Hu, Fu, Gorman, Johnson, Genereaux, Erdal, Gupta, Diaz-Pinto,
  Dourson, Maier-Hein, Jaeger, Baumgartner, Kalpathy-Cramer, Flores, Kirby,
  Cooper, Roth, Xu, Bericat, Floca, Zhou, Shuaib, Farahani, Maier-Hein,
  Aylward, Dogra, Ourselin and Feng}]{cardoso2022monai}
\bibinfo{author}{Cardoso, M.J.}, \bibinfo{author}{Li, W.},
  \bibinfo{author}{Brown, R.}, \bibinfo{author}{Ma, N.},
  \bibinfo{author}{Kerfoot, E.}, \bibinfo{author}{Wang, Y.},
  \bibinfo{author}{Murrey, B.}, \bibinfo{author}{Myronenko, A.},
  \bibinfo{author}{Zhao, C.}, \bibinfo{author}{Yang, D.},
  \bibinfo{author}{Nath, V.}, \bibinfo{author}{He, Y.}, \bibinfo{author}{Xu,
  Z.}, \bibinfo{author}{Hatamizadeh, A.}, \bibinfo{author}{Myronenko, A.},
  \bibinfo{author}{Zhu, W.}, \bibinfo{author}{Liu, Y.}, \bibinfo{author}{Zheng,
  M.}, \bibinfo{author}{Tang, Y.}, \bibinfo{author}{Yang, I.},
  \bibinfo{author}{Zephyr, M.}, \bibinfo{author}{Hashemian, B.},
  \bibinfo{author}{Alle, S.}, \bibinfo{author}{Darestani, M.Z.},
  \bibinfo{author}{Budd, C.}, \bibinfo{author}{Modat, M.},
  \bibinfo{author}{Vercauteren, T.}, \bibinfo{author}{Wang, G.},
  \bibinfo{author}{Li, Y.}, \bibinfo{author}{Hu, Y.}, \bibinfo{author}{Fu, Y.},
  \bibinfo{author}{Gorman, B.}, \bibinfo{author}{Johnson, H.},
  \bibinfo{author}{Genereaux, B.}, \bibinfo{author}{Erdal, B.S.},
  \bibinfo{author}{Gupta, V.}, \bibinfo{author}{Diaz-Pinto, A.},
  \bibinfo{author}{Dourson, A.}, \bibinfo{author}{Maier-Hein, L.},
  \bibinfo{author}{Jaeger, P.F.}, \bibinfo{author}{Baumgartner, M.},
  \bibinfo{author}{Kalpathy-Cramer, J.}, \bibinfo{author}{Flores, M.},
  \bibinfo{author}{Kirby, J.}, \bibinfo{author}{Cooper, L.A.D.},
  \bibinfo{author}{Roth, H.R.}, \bibinfo{author}{Xu, D.},
  \bibinfo{author}{Bericat, D.}, \bibinfo{author}{Floca, R.},
  \bibinfo{author}{Zhou, S.K.}, \bibinfo{author}{Shuaib, H.},
  \bibinfo{author}{Farahani, K.}, \bibinfo{author}{Maier-Hein, K.H.},
  \bibinfo{author}{Aylward, S.}, \bibinfo{author}{Dogra, P.},
  \bibinfo{author}{Ourselin, S.}, \bibinfo{author}{Feng, A.},
  \bibinfo{year}{2022}.
\newblock \bibinfo{title}{Monai: An open-source framework for deep learning in
  healthcare}.
\bibitem[{Caulo et~al.(2007)Caulo, Briganti, Mattei, Perfetti, Ferretti,
  Romani, Tartaro and Colosimo}]{HandKnob_Variability}
\bibinfo{author}{Caulo, M.}, \bibinfo{author}{Briganti, C.},
  \bibinfo{author}{Mattei, P.}, \bibinfo{author}{Perfetti, B.},
  \bibinfo{author}{Ferretti, A.}, \bibinfo{author}{Romani, G.},
  \bibinfo{author}{Tartaro, A.}, \bibinfo{author}{Colosimo, C.},
  \bibinfo{year}{2007}.
\newblock \bibinfo{title}{New morphologic variants of the hand motor cortex as
  seen with {MR} imaging in a large study population}.
\newblock \bibinfo{journal}{American Journal of Neuroradiology}
  \bibinfo{volume}{28}, \bibinfo{pages}{1480--1485}.
\newblock \DOIprefix\doi{10.3174/ajnr.a0597}.
\bibitem[{Chen et~al.(2020)Chen, Kornblith, Norouzi and Hinton}]{SimCLRMainRef}
\bibinfo{author}{Chen, T.}, \bibinfo{author}{Kornblith, S.},
  \bibinfo{author}{Norouzi, M.}, \bibinfo{author}{Hinton, G.},
  \bibinfo{year}{2020}.
\newblock \bibinfo{title}{A simple framework for contrastive learning of visual
  representations}, in: \bibinfo{booktitle}{International conference on machine
  learning}, \bibinfo{organization}{PMLR}. pp. \bibinfo{pages}{1597--1607}.
\bibitem[{Chlap et~al.(2021)Chlap, Min, Vandenberg, Dowling, Holloway and
  Haworth}]{DataAugmentMedDL}
\bibinfo{author}{Chlap, P.}, \bibinfo{author}{Min, H.},
  \bibinfo{author}{Vandenberg, N.}, \bibinfo{author}{Dowling, J.},
  \bibinfo{author}{Holloway, L.}, \bibinfo{author}{Haworth, A.},
  \bibinfo{year}{2021}.
\newblock \bibinfo{title}{A review of medical image data augmentation
  techniques for deep learning applications}.
\newblock \bibinfo{journal}{Journal of Medical Imaging and Radiation Oncology}
  \bibinfo{volume}{65}, \bibinfo{pages}{545--563}.
\newblock \DOIprefix\doi{https://doi.org/10.1111/1754-9485.13261}.
\bibitem[{{\c{C}}i{\c{c}}ek et~al.(2016){\c{C}}i{\c{c}}ek, Abdulkadir,
  Lienkamp, Brox and Ronneberger}]{Cisek3D_Unet}
\bibinfo{author}{{\c{C}}i{\c{c}}ek, {\"O}.}, \bibinfo{author}{Abdulkadir, A.},
  \bibinfo{author}{Lienkamp, S.S.}, \bibinfo{author}{Brox, T.},
  \bibinfo{author}{Ronneberger, O.}, \bibinfo{year}{2016}.
\newblock \bibinfo{title}{3d u-net: learning dense volumetric segmentation from
  sparse annotation}, in: \bibinfo{booktitle}{Medical Image Computing and
  Computer-Assisted Intervention--MICCAI 2016: 19th International Conference,
  Athens, Greece, October 17-21, 2016, Proceedings, Part II 19},
  \bibinfo{organization}{Springer}. pp. \bibinfo{pages}{424--432}.
\bibitem[{Clarisse et~al.(1997)Clarisse, Pertuzon, Ayachi, Francke
  et~al.}]{SulcalTopology}
\bibinfo{author}{Clarisse, J.}, \bibinfo{author}{Pertuzon, B.},
  \bibinfo{author}{Ayachi, M.}, \bibinfo{author}{Francke, J.}, et~al.,
  \bibinfo{year}{1997}.
\newblock \bibinfo{title}{Identification of the central sulcus using the
  scanner and mri}.
\newblock \bibinfo{journal}{Journal of Neuroradiology= Journal de
  Neuroradiologie} \bibinfo{volume}{24}, \bibinfo{pages}{187--204}.
\bibitem[{Collins et~al.(1994)Collins, Neelin, Peters and Evans}]{MNITemplate}
\bibinfo{author}{Collins, D.L.}, \bibinfo{author}{Neelin, P.},
  \bibinfo{author}{Peters, T.M.}, \bibinfo{author}{Evans, A.C.},
  \bibinfo{year}{1994}.
\newblock \bibinfo{title}{Automatic 3d intersubject registration of mr
  volumetric data in standardized talairach space.}
\newblock \bibinfo{journal}{Journal of computer assisted tomography}
  \bibinfo{volume}{18}, \bibinfo{pages}{192--205}.
\bibitem[{Desikan et~al.(2006)Desikan, S{\'{e}}gonne, Fischl, Quinn, Dickerson,
  Blacker, Buckner, Dale, Maguire, Hyman, Albert and Killiany}]{RegSulciSegm1}
\bibinfo{author}{Desikan, R.S.}, \bibinfo{author}{S{\'{e}}gonne, F.},
  \bibinfo{author}{Fischl, B.}, \bibinfo{author}{Quinn, B.T.},
  \bibinfo{author}{Dickerson, B.C.}, \bibinfo{author}{Blacker, D.},
  \bibinfo{author}{Buckner, R.L.}, \bibinfo{author}{Dale, A.M.},
  \bibinfo{author}{Maguire, R.P.}, \bibinfo{author}{Hyman, B.T.},
  \bibinfo{author}{Albert, M.S.}, \bibinfo{author}{Killiany, R.J.},
  \bibinfo{year}{2006}.
\newblock \bibinfo{title}{An automated labeling system for subdividing the
  human cerebral cortex on {MRI} scans into gyral based regions of interest}.
\newblock \bibinfo{journal}{{NeuroImage}} \bibinfo{volume}{31},
  \bibinfo{pages}{968--980}.
\newblock \DOIprefix\doi{10.1016/j.neuroimage.2006.01.021}.
\bibitem[{Dominic et~al.(2023)Dominic, Bhaskhar, Desai, Schmidt, Rubin, Gunel,
  Gold, Hargreaves, Lenchik, Boutin and Chaudhari}]{SimCLRUNET1}
\bibinfo{author}{Dominic, J.}, \bibinfo{author}{Bhaskhar, N.},
  \bibinfo{author}{Desai, A.D.}, \bibinfo{author}{Schmidt, A.},
  \bibinfo{author}{Rubin, E.}, \bibinfo{author}{Gunel, B.},
  \bibinfo{author}{Gold, G.E.}, \bibinfo{author}{Hargreaves, B.A.},
  \bibinfo{author}{Lenchik, L.}, \bibinfo{author}{Boutin, R.},
  \bibinfo{author}{Chaudhari, A.S.}, \bibinfo{year}{2023}.
\newblock \bibinfo{title}{Improving data-efficiency and robustness of medical
  imaging segmentation using inpainting-based self-supervised learning}.
\newblock \bibinfo{journal}{Bioengineering} \bibinfo{volume}{10}.
\newblock \DOIprefix\doi{10.3390/bioengineering10020207}.
\bibitem[{Ferrari et~al.(2016)Ferrari, Stockings, Khoo, Erskine, Degenhardt,
  Vos and Whiteford}]{BP_Burden}
\bibinfo{author}{Ferrari, A.J.}, \bibinfo{author}{Stockings, E.},
  \bibinfo{author}{Khoo, J.P.}, \bibinfo{author}{Erskine, H.E.},
  \bibinfo{author}{Degenhardt, L.}, \bibinfo{author}{Vos, T.},
  \bibinfo{author}{Whiteford, H.A.}, \bibinfo{year}{2016}.
\newblock \bibinfo{title}{The prevalence and burden of bipolar disorder:
  findings from the global burden of disease study 2013}.
\newblock \bibinfo{journal}{Bipolar Disorders} \bibinfo{volume}{18},
  \bibinfo{pages}{440--450}.
\newblock \DOIprefix\doi{10.1111/bdi.12423}.
\bibitem[{Fischl et~al.(1999)Fischl, Sereno and Dale}]{FS_Inflation}
\bibinfo{author}{Fischl, B.}, \bibinfo{author}{Sereno, M.I.},
  \bibinfo{author}{Dale, A.}, \bibinfo{year}{1999}.
\newblock \bibinfo{title}{Cortical surface-based analysis: Ii: Inflation,
  flattening, and a surface-based coordinate system}.
\newblock \bibinfo{journal}{NeuroImage} \bibinfo{volume}{9},
  \bibinfo{pages}{195 -- 207}.
\bibitem[{Gao et~al.(2020)Gao, Yoon, Wu and Chu}]{MultiTask1}
\bibinfo{author}{Gao, F.}, \bibinfo{author}{Yoon, H.}, \bibinfo{author}{Wu,
  T.}, \bibinfo{author}{Chu, X.}, \bibinfo{year}{2020}.
\newblock \bibinfo{title}{A feature transfer enabled multi-task deep learning
  model on medical imaging}.
\newblock \bibinfo{journal}{Expert Systems with Applications}
  \bibinfo{volume}{143}, \bibinfo{pages}{112957}.
\newblock \DOIprefix\doi{https://doi.org/10.1016/j.eswa.2019.112957}.
\bibitem[{Hesamian et~al.(2019)Hesamian, Jia, He and
  Kennedy}]{UnetMedSegmOverv}
\bibinfo{author}{Hesamian, M.H.}, \bibinfo{author}{Jia, W.},
  \bibinfo{author}{He, X.}, \bibinfo{author}{Kennedy, P.},
  \bibinfo{year}{2019}.
\newblock \bibinfo{title}{Deep learning techniques for medical image
  segmentation: Achievements and challenges}.
\newblock \bibinfo{journal}{Journal of Digital Imaging} \bibinfo{volume}{32},
  \bibinfo{pages}{582--596}.
\newblock \DOIprefix\doi{10.1007/s10278-019-00227-x}.
\bibitem[{Huang et~al.(2023)Huang, Pareek, Jensen, Lungren, Yeung and
  Chaudhari}]{SSTMedicalOverv}
\bibinfo{author}{Huang, S.C.}, \bibinfo{author}{Pareek, A.},
  \bibinfo{author}{Jensen, M.}, \bibinfo{author}{Lungren, M.P.},
  \bibinfo{author}{Yeung, S.}, \bibinfo{author}{Chaudhari, A.S.},
  \bibinfo{year}{2023}.
\newblock \bibinfo{title}{Self-supervised learning for medical image
  classification: a systematic review and implementation guidelines}.
\newblock \bibinfo{journal}{npj Digital Medicine} \bibinfo{volume}{6}.
\newblock \DOIprefix\doi{10.1038/s41746-023-00811-0}.
\bibitem[{Huntgeburth and Petrides(2012)}]{EarlyInterest2}
\bibinfo{author}{Huntgeburth, S.C.}, \bibinfo{author}{Petrides, M.},
  \bibinfo{year}{2012}.
\newblock \bibinfo{title}{Morphological patterns of the collateral sulcus in
  the human brain}.
\newblock \bibinfo{journal}{European Journal of Neuroscience}
  \bibinfo{volume}{35}, \bibinfo{pages}{1295--1311}.
\newblock \DOIprefix\doi{https://doi.org/10.1111/j.1460-9568.2012.08031.x}.
\bibitem[{Iglesias et~al.(2021)Iglesias, Billot, Balbastre, Tabari, Conklin,
  Gonz{\'a}lez, Alexander, Golland, Edlow, Fischl et~al.}]{SynthAugm2}
\bibinfo{author}{Iglesias, J.E.}, \bibinfo{author}{Billot, B.},
  \bibinfo{author}{Balbastre, Y.}, \bibinfo{author}{Tabari, A.},
  \bibinfo{author}{Conklin, J.}, \bibinfo{author}{Gonz{\'a}lez, R.G.},
  \bibinfo{author}{Alexander, D.C.}, \bibinfo{author}{Golland, P.},
  \bibinfo{author}{Edlow, B.L.}, \bibinfo{author}{Fischl, B.}, et~al.,
  \bibinfo{year}{2021}.
\newblock \bibinfo{title}{Joint super-resolution and synthesis of 1 mm
  isotropic mp-rage volumes from clinical mri exams with scans of different
  orientation, resolution and contrast}.
\newblock \bibinfo{journal}{Neuroimage} \bibinfo{volume}{237},
  \bibinfo{pages}{118206}.
\bibitem[{Jaiswal et~al.(2020)Jaiswal, Babu, Zadeh, Banerjee and
  Makedon}]{ContrSST_Revw}
\bibinfo{author}{Jaiswal, A.}, \bibinfo{author}{Babu, A.R.},
  \bibinfo{author}{Zadeh, M.Z.}, \bibinfo{author}{Banerjee, D.},
  \bibinfo{author}{Makedon, F.}, \bibinfo{year}{2020}.
\newblock \bibinfo{title}{A survey on contrastive self-supervised learning}.
\newblock \bibinfo{journal}{Technologies} \bibinfo{volume}{9},
  \bibinfo{pages}{2}.
\bibitem[{Jensen(2016)}]{Sulci_PhD}
\bibinfo{author}{Jensen, B.}, \bibinfo{year}{2016}.
\newblock \bibinfo{title}{Influence of Maturation, Pathology and Functional
  Lateralization on 3D Sulcal Morphology using MRI}.
\newblock Ph.D. thesis. Technical University of Denmark, DTU Compute.
\bibitem[{Kao et~al.(2007)Kao, Hofer, Sapiro, Stern, Rehm and
  Rottenberg}]{CurvDepth1}
\bibinfo{author}{Kao, C.Y.}, \bibinfo{author}{Hofer, M.},
  \bibinfo{author}{Sapiro, G.}, \bibinfo{author}{Stern, J.},
  \bibinfo{author}{Rehm, K.}, \bibinfo{author}{Rottenberg, D.A.},
  \bibinfo{year}{2007}.
\newblock \bibinfo{title}{A geometric method for automatic extraction of sulcal
  fundi}.
\newblock \bibinfo{journal}{IEEE transactions on medical imaging}
  \bibinfo{volume}{26}, \bibinfo{pages}{530--540}.
\bibitem[{Klein et~al.(2014)Klein, Rotarska-Jagiela, Genc, Sritharan, Mohr,
  Roux, Han, Kaiser, Singer and Uhlhaas}]{Gyrification_progression}
\bibinfo{author}{Klein, D.}, \bibinfo{author}{Rotarska-Jagiela, A.},
  \bibinfo{author}{Genc, E.}, \bibinfo{author}{Sritharan, S.},
  \bibinfo{author}{Mohr, H.}, \bibinfo{author}{Roux, F.}, \bibinfo{author}{Han,
  C.E.}, \bibinfo{author}{Kaiser, M.}, \bibinfo{author}{Singer, W.},
  \bibinfo{author}{Uhlhaas, P.J.}, \bibinfo{year}{2014}.
\newblock \bibinfo{title}{Adolescent brain maturation and cortical folding:
  Evidence for reductions in gyrification}.
\newblock \bibinfo{journal}{{PLoS} {ONE}} \bibinfo{volume}{9},
  \bibinfo{pages}{e84914}.
\newblock \DOIprefix\doi{10.1371/journal.pone.0084914}.
\bibitem[{Kochunov et~al.(2005)Kochunov, Mangin, Coyle, Lancaster, Thompson,
  Rivi{\`{e}}re, Cointepas, R{\'{e}}gis, Schlosser, Royall, Zilles, Mazziotta,
  Toga and Fox}]{NarrowSulci}
\bibinfo{author}{Kochunov, P.}, \bibinfo{author}{Mangin, J.F.},
  \bibinfo{author}{Coyle, T.}, \bibinfo{author}{Lancaster, J.},
  \bibinfo{author}{Thompson, P.}, \bibinfo{author}{Rivi{\`{e}}re, D.},
  \bibinfo{author}{Cointepas, Y.}, \bibinfo{author}{R{\'{e}}gis, J.},
  \bibinfo{author}{Schlosser, A.}, \bibinfo{author}{Royall, D.R.},
  \bibinfo{author}{Zilles, K.}, \bibinfo{author}{Mazziotta, J.},
  \bibinfo{author}{Toga, A.}, \bibinfo{author}{Fox, P.T.},
  \bibinfo{year}{2005}.
\newblock \bibinfo{title}{Age-related morphology trends of cortical sulci}.
\newblock \bibinfo{journal}{Human Brain Mapping} \bibinfo{volume}{26},
  \bibinfo{pages}{210--220}.
\newblock \URLprefix \url{https://doi.org/10.1002/hbm.20198},
  \DOIprefix\doi{10.1002/hbm.20198}.
\bibitem[{Kochunov et~al.(2011)Kochunov, Rogers, Mangin and
  Lancaster}]{BvisaOvervewPipeline}
\bibinfo{author}{Kochunov, P.}, \bibinfo{author}{Rogers, W.},
  \bibinfo{author}{Mangin, J.F.}, \bibinfo{author}{Lancaster, J.},
  \bibinfo{year}{2011}.
\newblock \bibinfo{title}{A library of cortical morphology analysis tools to
  study development, aging and genetics of cerebral cortex}.
\newblock \bibinfo{journal}{Neuroinformatics} \bibinfo{volume}{10},
  \bibinfo{pages}{81--96}.
\newblock \DOIprefix\doi{10.1007/s12021-011-9127-9}.
\bibitem[{Leroy et~al.(2015)Leroy, Cai, Bogart, Dubois, Coulon, Monzalvo,
  Fischer, Glasel, der Haegen, Bénézit, Lin, Kennedy, Ihara, Hertz-Pannier,
  Moutard, Poupon, Brysbaert, Roberts, Hopkins, Mangin and
  Dehaene-Lambertz}]{BVisaRef2}
\bibinfo{author}{Leroy, F.}, \bibinfo{author}{Cai, Q.},
  \bibinfo{author}{Bogart, S.L.}, \bibinfo{author}{Dubois, J.},
  \bibinfo{author}{Coulon, O.}, \bibinfo{author}{Monzalvo, K.},
  \bibinfo{author}{Fischer, C.}, \bibinfo{author}{Glasel, H.},
  \bibinfo{author}{der Haegen, L.V.}, \bibinfo{author}{Bénézit, A.},
  \bibinfo{author}{Lin, C.P.}, \bibinfo{author}{Kennedy, D.N.},
  \bibinfo{author}{Ihara, A.S.}, \bibinfo{author}{Hertz-Pannier, L.},
  \bibinfo{author}{Moutard, M.L.}, \bibinfo{author}{Poupon, C.},
  \bibinfo{author}{Brysbaert, M.}, \bibinfo{author}{Roberts, N.},
  \bibinfo{author}{Hopkins, W.D.}, \bibinfo{author}{Mangin, J.F.},
  \bibinfo{author}{Dehaene-Lambertz, G.}, \bibinfo{year}{2015}.
\newblock \bibinfo{title}{New human-specific brain landmark: The depth
  asymmetry of superior temporal sulcus}.
\newblock \bibinfo{journal}{Proceedings of the National Academy of Sciences}
  \bibinfo{volume}{112}, \bibinfo{pages}{1208--1213}.
\newblock \DOIprefix\doi{10.1073/pnas.1412389112}.
\bibitem[{Lin et~al.(2021a)Lin, Ye and Zhang}]{WeightedMT}
\bibinfo{author}{Lin, B.}, \bibinfo{author}{Ye, F.}, \bibinfo{author}{Zhang,
  Y.}, \bibinfo{year}{2021}a.
\newblock \bibinfo{title}{A closer look at loss weighting in multi-task
  learning}.
\newblock \bibinfo{journal}{arXiv preprint arXiv:2111.10603} .
\bibitem[{Lin et~al.(2021b)Lin, Huang, Chou, Yang, Lo, Tsai and
  Lin}]{GyrificationAging}
\bibinfo{author}{Lin, H.Y.}, \bibinfo{author}{Huang, C.C.},
  \bibinfo{author}{Chou, K.H.}, \bibinfo{author}{Yang, A.C.},
  \bibinfo{author}{Lo, C.Y.Z.}, \bibinfo{author}{Tsai, S.J.},
  \bibinfo{author}{Lin, C.P.}, \bibinfo{year}{2021}b.
\newblock \bibinfo{title}{Differential patterns of gyral and sulcal
  morphological changes during normal aging process}.
\newblock \bibinfo{journal}{Frontiers in Aging Neuroscience}
  \bibinfo{volume}{13}.
\newblock \DOIprefix\doi{10.3389/fnagi.2021.625931}.
\bibitem[{Liu et~al.(2021)Liu, Song, Liu and Zhang}]{DL_MedSegm}
\bibinfo{author}{Liu, X.}, \bibinfo{author}{Song, L.}, \bibinfo{author}{Liu,
  S.}, \bibinfo{author}{Zhang, Y.}, \bibinfo{year}{2021}.
\newblock \bibinfo{title}{A review of deep-learning-based medical image
  segmentation methods}.
\newblock \bibinfo{journal}{Sustainability} \bibinfo{volume}{13},
  \bibinfo{pages}{1224}.
\newblock \DOIprefix\doi{10.3390/su13031224}.
\bibitem[{Lyu et~al.(2021)Lyu, Bao, Hao, Yao, Miller, Voorhies, Taylor, Bunge,
  Weiner and Landman}]{SphericalCNNPFCS}
\bibinfo{author}{Lyu, I.}, \bibinfo{author}{Bao, S.}, \bibinfo{author}{Hao,
  L.}, \bibinfo{author}{Yao, J.}, \bibinfo{author}{Miller, J.A.},
  \bibinfo{author}{Voorhies, W.}, \bibinfo{author}{Taylor, W.D.},
  \bibinfo{author}{Bunge, S.A.}, \bibinfo{author}{Weiner, K.S.},
  \bibinfo{author}{Landman, B.A.}, \bibinfo{year}{2021}.
\newblock \bibinfo{title}{Labeling lateral prefrontal sulci using spherical
  data augmentation and context-aware training}.
\newblock \bibinfo{journal}{NeuroImage} \bibinfo{volume}{229},
  \bibinfo{pages}{117758}.
\newblock \DOIprefix\doi{https://doi.org/10.1016/j.neuroimage.2021.117758}.
\bibitem[{van~der Maaten and Hinton(2008)}]{TSNE}
\bibinfo{author}{van~der Maaten, L.}, \bibinfo{author}{Hinton, G.},
  \bibinfo{year}{2008}.
\newblock \bibinfo{title}{Visualizing data using t-sne}.
\newblock \bibinfo{journal}{Journal of Machine Learning Research}
  \bibinfo{volume}{9}, \bibinfo{pages}{2579--2605}.
\newblock \URLprefix \url{http://jmlr.org/papers/v9/vandermaaten08a.html}.
\bibitem[{Mangin et~al.(1995)Mangin, Frouin, Bloch, Rogis and
  Lopez-Krahe}]{EarlySInterest}
\bibinfo{author}{Mangin, J.F.}, \bibinfo{author}{Frouin, V.},
  \bibinfo{author}{Bloch, I.}, \bibinfo{author}{Rogis, J.},
  \bibinfo{author}{Lopez-Krahe, J.}, \bibinfo{year}{1995}.
\newblock \bibinfo{title}{From 3d magnetic resonance images to structural
  representations of the cortex topography using topology preserving
  deformations}.
\newblock \bibinfo{journal}{Journal of Mathematical Imaging and Vision}
  \bibinfo{volume}{5}, \bibinfo{pages}{297--318}.
\newblock \DOIprefix\doi{10.1007/bf01250286}.
\bibitem[{McConnell(1995)}]{FreeSuerferCortexMeshing}
\bibinfo{author}{McConnell, S.K.}, \bibinfo{year}{1995}.
\newblock \bibinfo{title}{Constructing the cerebral cortex: neurogenesis and
  fate determination}.
\newblock \bibinfo{journal}{Neuron} \bibinfo{volume}{15},
  \bibinfo{pages}{761--768}.
\bibitem[{Millier et~al.(2014)Millier, Schmidt, Angermeyer, Chauhan, Murthy,
  Toumi and Cadi-Soussi}]{SZ_burden}
\bibinfo{author}{Millier, A.}, \bibinfo{author}{Schmidt, U.},
  \bibinfo{author}{Angermeyer, M.}, \bibinfo{author}{Chauhan, D.},
  \bibinfo{author}{Murthy, V.}, \bibinfo{author}{Toumi, M.},
  \bibinfo{author}{Cadi-Soussi, N.}, \bibinfo{year}{2014}.
\newblock \bibinfo{title}{Humanistic burden in schizophrenia: A literature
  review}.
\newblock \bibinfo{journal}{Journal of Psychiatric Research}
  \bibinfo{volume}{54}, \bibinfo{pages}{85--93}.
\newblock \DOIprefix\doi{10.1016/j.jpsychires.2014.03.021}.
\bibitem[{Ochiai et~al.(2004)Ochiai, Grimault, Scavarda, Roch, Hori, Rivière,
  Mangin and Régis}]{BVisaRef1}
\bibinfo{author}{Ochiai, T.}, \bibinfo{author}{Grimault, S.},
  \bibinfo{author}{Scavarda, D.}, \bibinfo{author}{Roch, G.},
  \bibinfo{author}{Hori, T.}, \bibinfo{author}{Rivière, D.},
  \bibinfo{author}{Mangin, J.F.}, \bibinfo{author}{Régis, J.},
  \bibinfo{year}{2004}.
\newblock \bibinfo{title}{Sulcal pattern and morphology of the superior
  temporal sulcus}.
\newblock \bibinfo{journal}{NeuroImage} \bibinfo{volume}{22},
  \bibinfo{pages}{706--719}.
\newblock \DOIprefix\doi{https://doi.org/10.1016/j.neuroimage.2004.01.023}.
\bibitem[{Oord et~al.(2018)Oord, Li and Vinyals}]{InfoNCE}
\bibinfo{author}{Oord, A.v.d.}, \bibinfo{author}{Li, Y.},
  \bibinfo{author}{Vinyals, O.}, \bibinfo{year}{2018}.
\newblock \bibinfo{title}{Representation learning with contrastive predictive
  coding}.
\newblock \bibinfo{journal}{arXiv preprint arXiv:1807.03748} .
\bibitem[{Perrot et~al.(2011)Perrot, Rivière and Mangin}]{BvisaOldPaper}
\bibinfo{author}{Perrot, M.}, \bibinfo{author}{Rivière, D.},
  \bibinfo{author}{Mangin, J.F.}, \bibinfo{year}{2011}.
\newblock \bibinfo{title}{Cortical sulci recognition and spatial
  normalization}.
\newblock \bibinfo{journal}{Medical Image Analysis} \bibinfo{volume}{15},
  \bibinfo{pages}{529--550}.
\newblock \DOIprefix\doi{https://doi.org/10.1016/j.media.2011.02.008}.
  \bibinfo{note}{special section on IPMI 2009}.
\bibitem[{Puonti et~al.(2016)Puonti, Iglesias and {Van Leemput}}]{SamSeg}
\bibinfo{author}{Puonti, O.}, \bibinfo{author}{Iglesias, J.E.},
  \bibinfo{author}{{Van Leemput}, K.}, \bibinfo{year}{2016}.
\newblock \bibinfo{title}{Fast and sequence-adaptive whole-brain segmentation
  using parametric bayesian modeling}.
\newblock \bibinfo{journal}{NeuroImage} \bibinfo{volume}{143},
  \bibinfo{pages}{235--249}.
\newblock \DOIprefix\doi{https://doi.org/10.1016/j.neuroimage.2016.09.011}.
\bibitem[{Robinson and Bergen(2021)}]{SZ_BP_prevelance}
\bibinfo{author}{Robinson, N.}, \bibinfo{author}{Bergen, S.E.},
  \bibinfo{year}{2021}.
\newblock \bibinfo{title}{Environmental risk factors for schizophrenia and
  bipolar disorder and their relationship to genetic risk: Current knowledge
  and future directions}.
\newblock \bibinfo{journal}{Frontiers in Genetics} \bibinfo{volume}{12}.
\newblock \DOIprefix\doi{10.3389/fgene.2021.686666}.
\bibitem[{Roell et~al.(2021)Roell, Cachia, Matejko, Houdé, Ansari and
  Borst}]{BVisaRef3}
\bibinfo{author}{Roell, M.}, \bibinfo{author}{Cachia, A.},
  \bibinfo{author}{Matejko, A.}, \bibinfo{author}{Houdé, O.},
  \bibinfo{author}{Ansari, D.}, \bibinfo{author}{Borst, G.},
  \bibinfo{year}{2021}.
\newblock \bibinfo{title}{Sulcation of the intraparietal sulcus is related to
  symbolic but not non-symbolic number skills}.
\newblock \bibinfo{journal}{Developmental Cognitive Neuroscience}
  \bibinfo{volume}{51}, \bibinfo{pages}{100998}.
\newblock \DOIprefix\doi{https://doi.org/10.1016/j.dcn.2021.100998}.
\bibitem[{Rohlfing et~al.(2004)Rohlfing, Brandt, Menzel and Maurer~Jr}]{MAS}
\bibinfo{author}{Rohlfing, T.}, \bibinfo{author}{Brandt, R.},
  \bibinfo{author}{Menzel, R.}, \bibinfo{author}{Maurer~Jr, C.R.},
  \bibinfo{year}{2004}.
\newblock \bibinfo{title}{Evaluation of atlas selection strategies for
  atlas-based image segmentation with application to confocal microscopy images
  of bee brains}.
\newblock \bibinfo{journal}{NeuroImage} \bibinfo{volume}{21},
  \bibinfo{pages}{1428--1442}.
\bibitem[{Salehi et~al.(2017)Salehi, Erdogmus and Gholipour}]{TverskyLoss}
\bibinfo{author}{Salehi, S.S.M.}, \bibinfo{author}{Erdogmus, D.},
  \bibinfo{author}{Gholipour, A.}, \bibinfo{year}{2017}.
\newblock \bibinfo{title}{Tversky loss function for image segmentation using 3d
  fully convolutional deep networks}, in: \bibinfo{editor}{Wang, Q.},
  \bibinfo{editor}{Shi, Y.}, \bibinfo{editor}{Suk, H.I.},
  \bibinfo{editor}{Suzuki, K.} (Eds.), \bibinfo{booktitle}{Machine Learning in
  Medical Imaging}, \bibinfo{publisher}{Springer International Publishing},
  \bibinfo{address}{Cham}. pp. \bibinfo{pages}{379--387}.
\bibitem[{Schindler and Dellaert(2004)}]{EM_Paper}
\bibinfo{author}{Schindler, G.}, \bibinfo{author}{Dellaert, F.},
  \bibinfo{year}{2004}.
\newblock \bibinfo{title}{Atlanta world: An expectation maximization framework
  for simultaneous low-level edge grouping and camera calibration in complex
  man-made environments}, in: \bibinfo{booktitle}{Proceedings of the 2004 IEEE
  Computer Society Conference on Computer Vision and Pattern Recognition, 2004.
  CVPR 2004.}, \bibinfo{organization}{IEEE}. pp. \bibinfo{pages}{I--I}.
\bibitem[{Shi et~al.(2007)Shi, Tu, Reiss, Dutton, Lee, Galaburda, Dinov,
  Thompson and Toga}]{CurvDepth3}
\bibinfo{author}{Shi, Y.}, \bibinfo{author}{Tu, Z.}, \bibinfo{author}{Reiss,
  A.L.}, \bibinfo{author}{Dutton, R.A.}, \bibinfo{author}{Lee, A.D.},
  \bibinfo{author}{Galaburda, A.M.}, \bibinfo{author}{Dinov, I.},
  \bibinfo{author}{Thompson, P.M.}, \bibinfo{author}{Toga, A.W.},
  \bibinfo{year}{2007}.
\newblock \bibinfo{title}{Joint sulci detection using graphical models and
  boosted priors}, in: \bibinfo{booktitle}{IPMI}, pp. \bibinfo{pages}{98--109}.
\bibitem[{Thorup et~al.(2018)Thorup, Hemager, S{\o}ndergaard, Gregersen,
  Pr{\o}sch, Krantz, Brandt, Carmichael, Melau, Ellersgaard, Burton, Greve,
  Uddin, Ohland, Nejad, Johnsen, van Themaat, Andreassen, Vedum, Knudsen,
  Stadsgaard, Jepsen, Siebner, {\O}stergaard, Bliksted, Plessen, Mors and
  Nordentoft}]{VIA11_Overview}
\bibinfo{author}{Thorup, A.A.E.}, \bibinfo{author}{Hemager, N.},
  \bibinfo{author}{S{\o}ndergaard, A.}, \bibinfo{author}{Gregersen, M.},
  \bibinfo{author}{Pr{\o}sch, {\AA}.K.}, \bibinfo{author}{Krantz, M.F.},
  \bibinfo{author}{Brandt, J.M.}, \bibinfo{author}{Carmichael, L.},
  \bibinfo{author}{Melau, M.}, \bibinfo{author}{Ellersgaard, D.V.},
  \bibinfo{author}{Burton, B.K.}, \bibinfo{author}{Greve, A.N.},
  \bibinfo{author}{Uddin, M.J.}, \bibinfo{author}{Ohland, J.},
  \bibinfo{author}{Nejad, A.B.}, \bibinfo{author}{Johnsen, L.K.},
  \bibinfo{author}{van Themaat, A.H.V.L.}, \bibinfo{author}{Andreassen, A.K.},
  \bibinfo{author}{Vedum, L.}, \bibinfo{author}{Knudsen, C.B.},
  \bibinfo{author}{Stadsgaard, H.}, \bibinfo{author}{Jepsen, J.R.M.},
  \bibinfo{author}{Siebner, H.R.}, \bibinfo{author}{{\O}stergaard, L.},
  \bibinfo{author}{Bliksted, V.F.}, \bibinfo{author}{Plessen, K.J.},
  \bibinfo{author}{Mors, O.}, \bibinfo{author}{Nordentoft, M.},
  \bibinfo{year}{2018}.
\newblock \bibinfo{title}{The danish high risk and resilience
  study{\textemdash}{VIA} 11: Study protocol for the first follow-up of the
  {VIA} 7 cohort -522 children born to parents with schizophrenia spectrum
  disorders or bipolar disorder and controls being re-examined for the first
  time at age 11}.
\newblock \bibinfo{journal}{Frontiers in Psychiatry} \bibinfo{volume}{9}.
\newblock \DOIprefix\doi{10.3389/fpsyt.2018.00661}.
\bibitem[{Thorup et~al.(2015)Thorup, Jepsen, Ellersgaard, Burton, Christiani,
  Hemager, Skj{\ae}rb{\ae}k, Ranning, Spang, Gantriis, Greve, Zahle, Mors,
  Plessen and Nordentoft}]{VIA7_Study_Protocol}
\bibinfo{author}{Thorup, A.A.E.}, \bibinfo{author}{Jepsen, J.R.},
  \bibinfo{author}{Ellersgaard, D.V.}, \bibinfo{author}{Burton, B.K.},
  \bibinfo{author}{Christiani, C.J.}, \bibinfo{author}{Hemager, N.},
  \bibinfo{author}{Skj{\ae}rb{\ae}k, M.}, \bibinfo{author}{Ranning, A.},
  \bibinfo{author}{Spang, K.S.}, \bibinfo{author}{Gantriis, D.L.},
  \bibinfo{author}{Greve, A.N.}, \bibinfo{author}{Zahle, K.K.},
  \bibinfo{author}{Mors, O.}, \bibinfo{author}{Plessen, K.J.},
  \bibinfo{author}{Nordentoft, M.}, \bibinfo{year}{2015}.
\newblock \bibinfo{title}{The danish high risk and resilience study
  {\textendash} {VIA} 7 - a cohort study of 520 7-year-old children born of
  parents diagnosed with either schizophrenia, bipolar disorder or neither of
  these two mental disorders}.
\newblock \bibinfo{journal}{{BMC} Psychiatry} \bibinfo{volume}{15}.
\newblock \DOIprefix\doi{10.1186/s12888-015-0616-5}.
\bibitem[{Vivodtzev et~al.(2003)Vivodtzev, Linsen, Bonneau, Hamann, Joy and
  Olshausen}]{CurvDepth2}
\bibinfo{author}{Vivodtzev, F.}, \bibinfo{author}{Linsen, L.},
  \bibinfo{author}{Bonneau, G.P.}, \bibinfo{author}{Hamann, B.},
  \bibinfo{author}{Joy, K.}, \bibinfo{author}{Olshausen, B.A.},
  \bibinfo{year}{2003}.
\newblock \bibinfo{title}{Hierachical isosurface segmentation based on discrete
  curvature}.
\newblock \bibinfo{journal}{UC Davis: Institute for Data Analysis and
  Visualization} .
\bibitem[{White et~al.(2010)White, Su, Schmidt, Kao and
  Sapiro}]{GyrificationDevelopm}
\bibinfo{author}{White, T.}, \bibinfo{author}{Su, S.},
  \bibinfo{author}{Schmidt, M.}, \bibinfo{author}{Kao, C.Y.},
  \bibinfo{author}{Sapiro, G.}, \bibinfo{year}{2010}.
\newblock \bibinfo{title}{The development of gyrification in childhood and
  adolescence}.
\newblock \bibinfo{journal}{Brain and Cognition} \bibinfo{volume}{72},
  \bibinfo{pages}{36--45}.
\newblock \DOIprefix\doi{10.1016/j.bandc.2009.10.009}.
\bibitem[{Willbrand et~al.(2022)Willbrand, Parker, Voorhies, Miller, Lyu,
  Hallock, Aponik-Gremillion, Koslov, Bunge, Foster and and}]{SpherUsePaperEx}
\bibinfo{author}{Willbrand, E.H.}, \bibinfo{author}{Parker, B.J.},
  \bibinfo{author}{Voorhies, W.I.}, \bibinfo{author}{Miller, J.A.},
  \bibinfo{author}{Lyu, I.}, \bibinfo{author}{Hallock, T.},
  \bibinfo{author}{Aponik-Gremillion, L.}, \bibinfo{author}{Koslov, S.R.},
  \bibinfo{author}{Bunge, S.A.}, \bibinfo{author}{Foster, B.L.},
  \bibinfo{author}{and, K.S.W.}, \bibinfo{year}{2022}.
\newblock \bibinfo{title}{Uncovering a tripartite landmark in posterior
  cingulate cortex}.
\newblock \bibinfo{journal}{Science Advances} \bibinfo{volume}{8}.
\newblock \DOIprefix\doi{10.1126/sciadv.abn9516}.
\bibitem[{Yang and Kruggel(2008)}]{OldSulcSegmReview}
\bibinfo{author}{Yang, F.}, \bibinfo{author}{Kruggel, F.},
  \bibinfo{year}{2008}.
\newblock \bibinfo{title}{Automatic segmentation of human brain sulci}.
\newblock \bibinfo{journal}{Medical Image Analysis} \bibinfo{volume}{12},
  \bibinfo{pages}{442--451}.
\newblock \DOIprefix\doi{https://doi.org/10.1016/j.media.2008.01.003}.
\bibitem[{Yang et~al.(2019)Yang, Wang, Rollins, Leming, Li{\`{o}}, Suckling,
  Murray, Garrison and Cachia}]{RegCNNParcSulc}
\bibinfo{author}{Yang, J.}, \bibinfo{author}{Wang, D.},
  \bibinfo{author}{Rollins, C.}, \bibinfo{author}{Leming, M.},
  \bibinfo{author}{Li{\`{o}}, P.}, \bibinfo{author}{Suckling, J.},
  \bibinfo{author}{Murray, G.}, \bibinfo{author}{Garrison, J.},
  \bibinfo{author}{Cachia, A.}, \bibinfo{year}{2019}.
\newblock \bibinfo{title}{Volumetric segmentation and characterisation of the
  paracingulate sulcus on {MRI} scans}.
\newblock \bibinfo{journal}{bioRxiv} \DOIprefix\doi{10.1101/859496}.
\bibitem[{Yaniv et~al.(2017)Yaniv, Lowekamp, Johnson and Beare}]{SimpleITK}
\bibinfo{author}{Yaniv, Z.}, \bibinfo{author}{Lowekamp, B.C.},
  \bibinfo{author}{Johnson, H.J.}, \bibinfo{author}{Beare, R.},
  \bibinfo{year}{2017}.
\newblock \bibinfo{title}{{SimpleITK} image-analysis notebooks: a collaborative
  environment for education and reproducible research}.
\newblock \bibinfo{journal}{Journal of Digital Imaging} \bibinfo{volume}{31},
  \bibinfo{pages}{290--303}.
\newblock \DOIprefix\doi{10.1007/s10278-017-0037-8}.
\bibitem[{Zeng et~al.(2021)Zeng, Kheir, Zeng and Shi}]{SimCLRUNET2}
\bibinfo{author}{Zeng, D.}, \bibinfo{author}{Kheir, J.N.},
  \bibinfo{author}{Zeng, P.}, \bibinfo{author}{Shi, Y.}, \bibinfo{year}{2021}.
\newblock \bibinfo{title}{Contrastive learning with temporal correlated medical
  images: A case study using lung segmentation in chest x-rays (invited
  paper)}, in: \bibinfo{booktitle}{2021 IEEE/ACM International Conference On
  Computer Aided Design (ICCAD)}, pp. \bibinfo{pages}{1--7}.
\bibitem[{Zhang et~al.(2020)Zhang, Wang, Gao, Li, Zhang, Lin, Hou, Yu, Wang and
  Liu}]{BV_SS_1}
\bibinfo{author}{Zhang, Z.}, \bibinfo{author}{Wang, Y.}, \bibinfo{author}{Gao,
  Y.}, \bibinfo{author}{Li, Z.}, \bibinfo{author}{Zhang, S.},
  \bibinfo{author}{Lin, X.}, \bibinfo{author}{Hou, Z.}, \bibinfo{author}{Yu,
  Q.}, \bibinfo{author}{Wang, X.}, \bibinfo{author}{Liu, S.},
  \bibinfo{year}{2020}.
\newblock \bibinfo{title}{Morphological changes in the central sulcus of
  children with isolated growth hormone deficiency versus idiopathic short
  stature}.
\newblock \bibinfo{journal}{Developmental Neurobiology} \bibinfo{volume}{81},
  \bibinfo{pages}{36--46}.
\newblock \DOIprefix\doi{10.1002/dneu.22797}.
\bibitem[{Zhou et~al.(2021)Zhou, Chen, Li, Liu, Xu, Wang, Yap and
  Shen}]{MultiTask2}
\bibinfo{author}{Zhou, Y.}, \bibinfo{author}{Chen, H.}, \bibinfo{author}{Li,
  Y.}, \bibinfo{author}{Liu, Q.}, \bibinfo{author}{Xu, X.},
  \bibinfo{author}{Wang, S.}, \bibinfo{author}{Yap, P.T.},
  \bibinfo{author}{Shen, D.}, \bibinfo{year}{2021}.
\newblock \bibinfo{title}{Multi-task learning for segmentation and
  classification of tumors in 3d automated breast ultrasound images}.
\newblock \bibinfo{journal}{Medical Image Analysis} \bibinfo{volume}{70},
  \bibinfo{pages}{101918}.
\newblock \DOIprefix\doi{https://doi.org/10.1016/j.media.2020.101918}.

\end{thebibliography}

\end{document}